\documentclass[10pt,twocolumn,letterpaper]{article}

\usepackage[pagenumbers]{cvpr} 

\usepackage{graphicx}
\usepackage{amsmath}
\usepackage{amssymb}
\usepackage{booktabs}
\usepackage{xcolor}

\usepackage[numbers]{natbib}
\usepackage[resetlabels]{multibib}
\newcites{supp}{References}

\usepackage[pagebackref,breaklinks,colorlinks,hypertexnames=false]{hyperref}

\newcommand{\nocontentsline}[3]{}
\newcommand{\tocless}[2]{\bgroup\let\addcontentsline=\nocontentsline#1{#2}\egroup}

\usepackage[capitalize]{cleveref}
\crefname{section}{Sec.}{Secs.}
\Crefname{section}{Section}{Sections}
\Crefname{table}{Table}{Tables}
\crefname{table}{Tab.}{Tabs.}

\usepackage{float}
\usepackage{bbm}
\usepackage{dutchcal}
\usepackage{multirow}

\begin{document}

\title{Box-Level Active Detection}

\author{Mengyao Lyu$^{1,2,3}$
\: Jundong Zhou$^{1,2,3}$
\: Hui Chen$^{1,2}$
\: Yijie Huang$^{4}$
\: Dongdong Yu$^{4}$\\
\: Yaqian Li$^{4}$
\: Yandong Guo$^{4}$
\: Yuchen Guo$^{1,2}$
\: Liuyu Xiang$^{5*}$
\: Guiguang Ding$^{1,2}\thanks{Corresponding Authors.}$\quad
\\
$^{1}$Tsinghua University\: $^{2}$BNRist\: $^{3}$Hangzhou Zhuoxi Institute of Brain and Intelligence\\
$^{4}$OPPO Research Institute\: $^{5}$Beijing University of Posts and Telecommunications\\
{\tt\small 
    \{mengyao.lyu,jundong.zhou\}@outlook.com\:
    \{huangyijie,yudongdong,liyaqian,guoyandong\}@oppo.com
}\\
{\tt\small 
    \{jichenhui2012,yuchen.w.guo\}@gmail.com\:
    xiangly@bupt.edu.cn\:
    dinggg@tsinghua.edu.cn\:
    }
}
\maketitle

\begin{abstract}
   Active learning selects informative samples for annotation within budget, which has proven efficient recently on object detection. 
   However, the widely used active detection benchmarks conduct \textbf{image-level evaluation}, which is unrealistic in human workload estimation and biased towards crowded images. 
   Furthermore, existing methods still perform \textbf{image-level annotation}, but equally scoring all targets within the same image incurs waste of budget and redundant labels.
   Having revealed above problems
   and limitations, we introduce a \textbf{box-level active detection} framework that controls a box-based budget per cycle, prioritizes informative targets and avoids redundancy for fair comparison and efficient application.
   
   Under the proposed box-level setting, we devise a novel pipeline, namely \textbf{Complementary Pseudo Active Strategy (ComPAS)}. It exploits both human annotations and the model intelligence in a complementary fashion: an efficient input-end committee queries labels for informative objects only; meantime well-learned targets are identified by the model and compensated with pseudo-labels.
   ComPAS consistently outperforms 10 competitors under 4 settings in a unified codebase. With supervision from labeled data only, it achieves 100\% supervised performance of VOC0712 with merely 19\% box annotations. On  the COCO dataset, it yields up to 4.3\% mAP improvement over the second-best method. 
   ComPAS also supports training with the unlabeled pool, where it surpasses 90\% COCO supervised performance with 85\% label reduction.
   Our source code is publicly available at \href{https://github.com/lyumengyao/blad}{https://github.com/lyumengyao/blad}.
\end{abstract}

\tocless\section{Introduction}
\label{sec:intro}

\begin{figure}[t]
  \centering
   \includegraphics[width=1\linewidth]{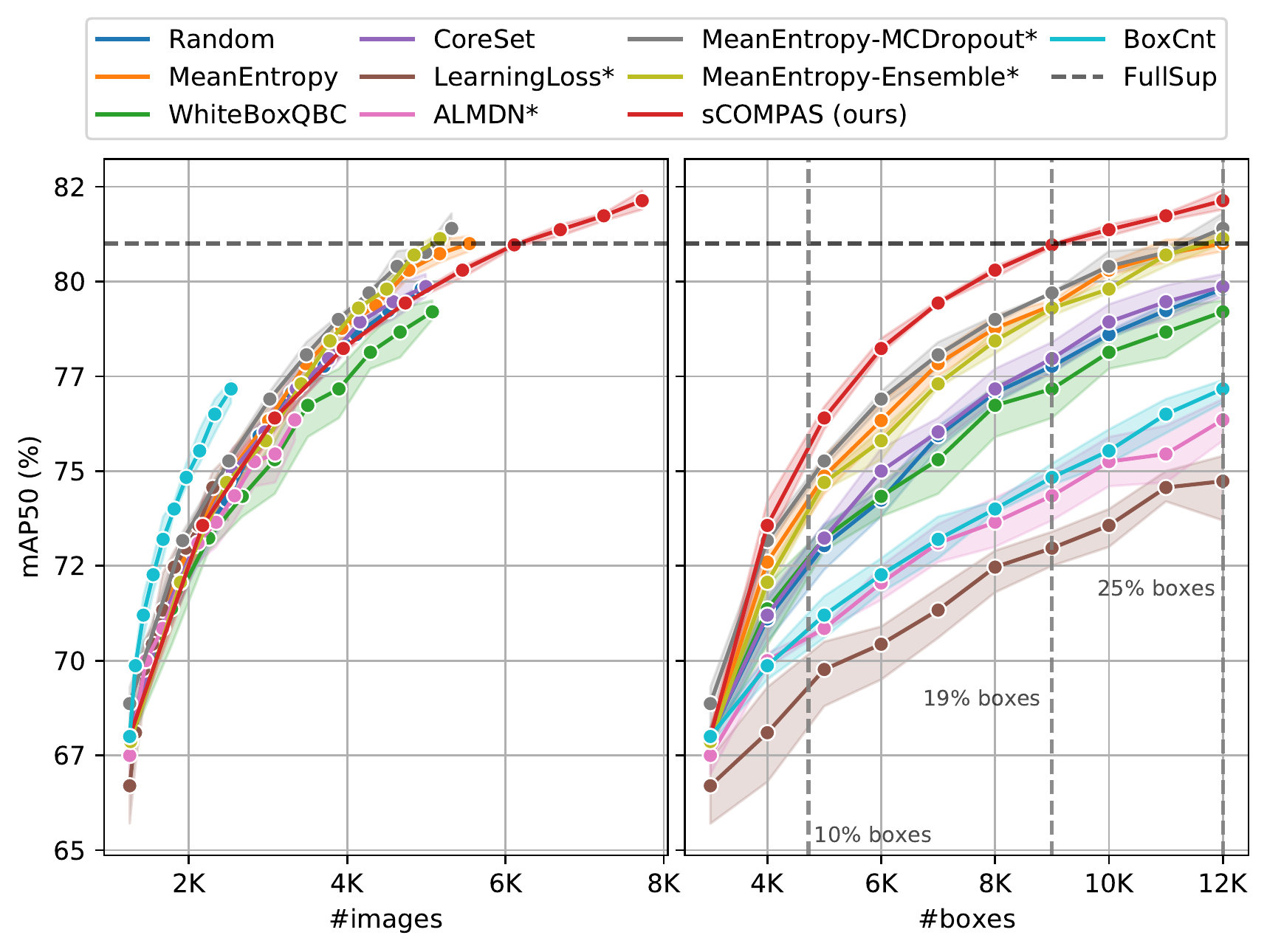}
   \caption{Active detection methods evaluated on VOC0712 under image-level (Left) and box-level (Right) settings. BoxCnt is our hack that simply queries potentially the most crowded images, which demonstrates that image-level evaluation is highly biased.
   Methods marked with * have specialized detector architectures.
   }
   \label{fig:image-level}
\end{figure}

Reducing the dependency on large-scale and well-annotated datasets for deep neural networks has received a growing interest in recent years, especially for the detection task, where the box-level annotation is highly demanding.
Among data-efficient training schemes, active detection methods~\cite{agarwal2020cdal,kao2018localizationaware,Desai2019adaptiveswitch,choi2021gmm,yuan2021miaod,mi2022ActiveTeacherSemiSupervised,elezi2022NotAllLabels,roy2018qbc} iterate over detector training, performance evaluation, informative image acquisition and human annotation. 
Despite recent progress, 
previous pool-based active detection methods still 
consider the subject of interest at the image-level: they conduct \textbf{image-level evaluation}, where the budget is controlled by the number of labeled images per cycle; afterwards, they perform exhaustive \textbf{image-level annotation}, where all instances of the same image are labeled. Such an image-level framework suffers from unfairness in model performance comparison and leads to a waste of annotation resources.

On the one hand, existing methods under the image-level evaluation assume equal budget for every image.
However, in real-world use cases, the workload of annotators is measured by bounding boxes~\cite{ren2020omniufo,gupta2019lvis}. 
As the image-level budget fails to reflect actual box-based costs, active detection methods are allowed to obscurely gain an advantage by querying box supervision as much as possible until the image-based budget is run out. 
In fact, according to our experiment shown in Fig.~\ref{fig:image-level}L, naively sampling potentially the most crowded images (dubbed as ``BoxCnt'') can surpass all elaborately designed methods, demonstrating the unfairness of image-level evaluation.
On the other hand, during human annotation, simply performing image-level exhaustive annotation is wasteful, since the informativeness of different targets involved in the same image can vary sharply. 
For example, a salient target of a common category might have been well-learned, whereas a distant or occluded variant could be more informative. As a result, annotating all instances amongst the same image as equals leads to a  waste of resources and redundant annotations (See Fig.~\ref{fig:violin}).

After revealing the above problems and limitations, we propose a new \textbf{box-level active detection} framework towards fair comparison and non-redundant human annotation.
For evaluation, our framework includes a more practical and unbiased criterion that controls the amount of queried boxes per cycle, enabling competing methods to be  assessed directly within realistic box-based budgets (as illustrated in Fig.~\ref{fig:image-level}R).
Considering the annotation, we advocate a box-level protocol that prioritizes top-ranked targets for annotation and discards well-learned counterparts to avoid redundancy.
Under the proposed framework, we develop a novel and efficient method namely \textbf{Com}plementary \textbf{P}seudo \textbf{A}ctive \textbf{S}trategy (\textbf{ComPAS}). It seamlessly integrates human efforts with model intelligence in actively acquiring informative targets via an \textit{input-end committee}, and meantime remedying the annotation of well-learned counterparts using \textit{online pseudo-labels}. 

In consideration of the active acquisition, 
concentrating resources on the most informative targets makes box-level informativeness estimation crucial.
Among active learning strategies, multi-model methods, such as Ensemble~\cite{beluch2018ensemble} and MCDropout~\cite{gal2017mcdropout}, have demonstrated superiority.
Built upon a model-end ensemble, query-by-committee methods select the most controversial candidates based on the voting of model members to minimize the version space~\cite{seung1992qbc,settles2012active}. 
However, directly adapting them to detection not only multiplies the computational cost in the committee construction, but complicates the detection hypothesis ensemble on the box-level.
Therefore, to harness the power of diversity without a heavy computational burden, orthogonal to model-end ensembles, we construct an input-end committee during the sampling stage. Variations are drawn from ubiquitous data augmentations and applied to unlabeled candidates, among which each perturbation can be considered as a \textit{cheap but effective committee} member towards version space minimization. 
When it comes to the box-level hypothesis ensemble, instead of performing pair-wise label assignment among all members~\cite{roy2018qbc}, we \textit{reduce the ensemble burden} by analyzing the disagreement between predictions of a reliable reference 
and other members.
Then the disagreement is quantified for both classification and localization to exploit the rich information in annotations.

Later during box-level annotation, the oracle only yields labels for challenging, controversial targets, leaving consistent ones unlabeled. 
Those unlabeled targets would be considered as the background class during the following training cycles, which severely harms the performance and poses a new challenge.
To compensate well-learned targets for missing annotation, we combine sparse ground truths with online pseudo-label generation, where in contrast to active sampling, confident model predictions are accepted as self-supervision signals.
The proposed box-level pipeline supports both labeled-only and mixed-supervision learning settings w/ or w/o the unlabeled image pool involved during training, which makes a fairer comparison with fully- and semi-supervised state-of-the-arts (SOTAs).

Our contributions can be summarized as follows: 
\begin{itemize}
    \itemsep-0.3em 
    \item We propose a box-level active detection framework, where we control box-based budgets for realistic and fair evaluation, and concentrate annotation resources on the most informative targets to avoid redundancy.
    \item We develop ComPAS, a novel method that seamlessly integrates model intelligence into human efforts via an input-end committee for challenging target annotation and pseudo-labeling for well-learned counterparts.
    \item We provide a unified codebase with implementations of active detection baselines and SOTAs, under which the superiority of ComPAS is demonstrated via extensive experiments.
\end{itemize}

\tocless\section{Related Work}

\begin{figure*}[t]
  \centering
   \includegraphics[width=1\linewidth]{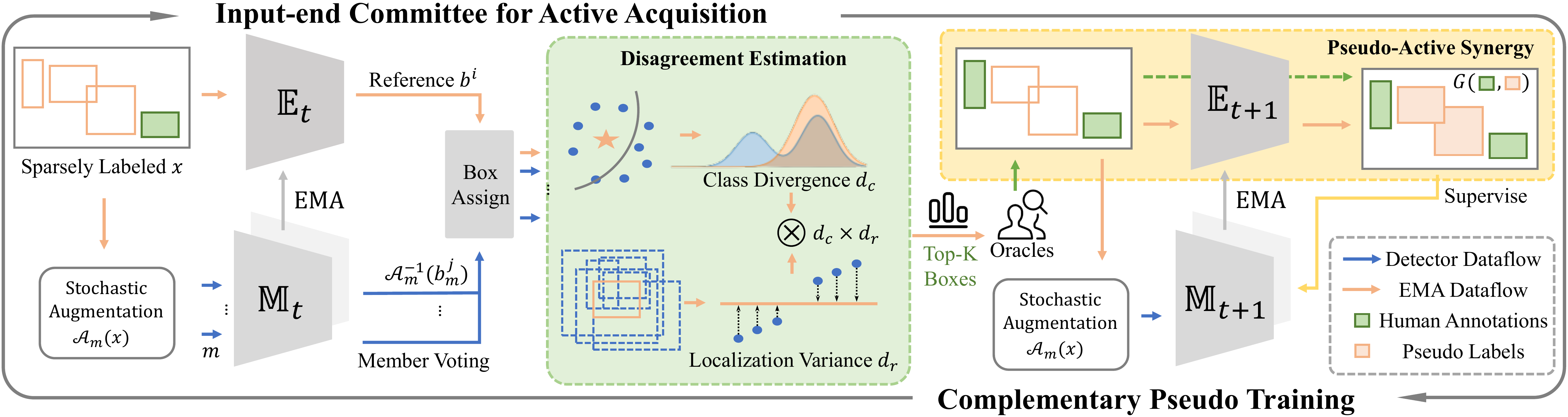}
   \caption{Overview of our ComPAS pipeline for box-level active detection, which iterates between active acquisition via the input-end committee and complementary training based on pseudo-active synergy. Only workflow of sparsely labeled images is shown for generality.}
   \label{fig:framework}
\end{figure*}
\textbf{Active scoring functions.}
Pool-based active detection strategies rely on scoring functions to rank sample candidates for annotation, which can be categorized into uncertainty-based~\cite{choi2021gmm,roy2018qbc,yuan2021miaod,yoo2019learningloss,hetao}, diversity-based~\cite{sener2018coreset,agarwal2020cdal} and hybrid methods~\cite{wu2022EntropyBasedActiveLearning,mi2022ActiveTeacherSemiSupervised}.
Uncertainty-based methods prioritize unconfident predictions based on posterior probability distributions~\cite{yuan2021miaod,gal2017mcdropout,beluch2018ensemble}, a specified loss prediction module~\cite{yoo2019learningloss}, or Gaussian mixture heads~\cite{choi2021gmm}.
To avoid sampling bias~\cite{dasgupta2011samplingbias,ren2020survey} in batch mode active learning induced by uncertainty, another strategy is to promote diversity for a more representative dataset, which is achieved by core-set selection~\cite{sener2018coreset,agarwal2020cdal}.
However, diversity-based methods incline to sample data points as far as possible
to cover the data manifold without considering density. Thus, {hybrid} methods that make trade-offs between diversity and uncertainty are proposed~\cite{wu2022EntropyBasedActiveLearning,mi2022ActiveTeacherSemiSupervised}. 
Besides adaption from classification-oriented methods, some recent research specially considers the localization subtask, of which the uncertainty is estimated via the inconsistency between RPN proposals and final predictions~\cite{kao2018localizationaware}, or a mixture density model~\cite{choi2021gmm}. However, they either impose limitations on the detector architecture, or require certain modifications to it, thus cannot be generalized. In contrast, our localization informativeness is efficiently estimated between stochastic perturbations of candidates without dependency on model architecture.

\textbf{Multi-model score ensemble.}
Based on the above active scoring functions, multi-model methods ensemble different hypotheses obtained via multiple training repetitions~\cite{beluch2018ensemble}, stochastic forward passes~\cite{gal2017mcdropout}, different model scales~\cite{roy2018qbc} 
or duplicated detection heads~\cite{choi2021gmm}.
Despite their effectiveness because of increased variety, ensemble methods have not been widely discussed in detection due to the computational burden and the box-level ensemble difficulty.
To reduce the computational cost,
some of the previous methods attempted at altering model inputs, such as image flipping~\cite{elezi2022NotAllLabels} and noise interfering~\cite{kao2018localizationaware}, but those variations are simple and limited.
During the result ensemble, unlike classification methods that can directly average over posterior distributions, existing adaptations towards detection mainly avoid the obstacles by image-level scoring followed by model-level aggregation~\cite{choi2021gmm}. However, when applied on the box-level, it requires pair-wise label assignment~\cite{roy2018qbc}, which further incurs computational cost.
In contrast, our input-based committee promotes diversity via stronger positional and color perturbations applied on more input members, and disagreement is efficiently analyzed between a reference and members.

\textbf{Implementation and evaluation.}
While most recent research on active detection is still evaluated on the image-level, our analysis has revealed that it is unrealistic and heavily biased. Furthermore, we suggest box-level annotation, which is attempted but neither well-explored \cite{Desai2020boxlevel} nor applicable to the in-domain task~\cite{qbox}. To this end, we present a strong pipeline that integrates both human annotations and machine predictions on the box-level.
We also note that previous active detection methods are compared without detector uniformity (\eg SSD~\cite{liu2016ssd}, Faster R-CNN~\cite{ren2015faster}, RetinaNet~\cite{lin2017retinafocal}), learning standardization (\eg runtime settings), benchmark consistency (VOC12/0712~\cite{everingham2010voc}, COCO2014/2017~\cite{lin2014microsoft}), and supervision differentiation (fully- or semi-supervised).
To help advance reproducible research, we introduce a shared implementation of methods based on the same detector,
train with similar procedures in a unified codebase, evaluate under the box-level criterion and support both labeled-only and mixed-supervision learning.

\tocless\section{ComPAS for Box-level Active Detection}
\label{sec:method}

\subsection{Problem Formulation}  
The pipeline of the box-level active detection is initialized at the active learning cycle $t=0$. A small set of images $\mathit{L}_0$ is randomly sampled and fully annotated with bounding boxes, whereas the majority of images are remained unlabeled $\mathit{U}_0$. Based on the current data pools, a generic object detector $\mathbb M(\theta_0)$ is obtained, evaluated and used for inferring on the unlabeled pool. Then a scoring function evaluates the informativeness of each unlabeled candidate and queries an oracle for labels. 
Different from image-level active detection methods that consider an image as the minimum annotation unit, we actively select top-ranked informative bounding box proposals for annotators to identify the objects of interest within the candidate regions. 
Such protocol actively prompts annotators with the potential boxes for correction, rather than leaving them passively spotting, marking and verifying all instances for every class among an image, which greatly helps narrow down the spatial search space and semantic options.

Since the first active sampling process $t \ge 1$, sparsely labeled images become available as $\mathit{S}_t$.
Then the detector ${\mathbb M(\theta_t)}$ is updated accordingly, with labeled images only ($\mathit{L}_t\cup \mathit{S}_t$), or with all images involved ($\mathit{L}_t\cup \mathit{S}_t \cup \mathit{U}_t$) in the mixed-supervision setting. 
In the subsequent active acquisition cycles, both sparsely labeled and unlabeled images are evaluated, among which top-ranked boxes with low overlap with existing ground truths are prompted for labels. This iteration repeats itself until the stopping criterion is reached. 

As illustrated in Fig.~\ref{fig:framework}, under the box-level active detection scenario, we propose a Complementary Pseudo Active Strategy (ComPAS), where the synergy between hard ground truth mining during active sampling (Sec.~\ref{sec:active}) and easy pseudo-label generation (Sec.~\ref{sec:training}) is exploited.

\subsection{Active Acquisition via Input-end Committee}
\label{sec:active}
Ensemble-based active learning has proven effective for classification~\cite{gal2017mcdropout,beluch2018ensemble} as well as detection under the box-level evaluation (shown in Fig.~\ref{fig:image-level}). The ensemble is also dubbed as a \textit{committee}~\cite{seung1992qbc,roy2018qbc} when the disagreement amongst \textit{member} hypotheses is estimated.
However, existing ensemble strategies mainly rely on model parameter duplication, referred to as model-end diversity, which induces extra computational cost.
Furthermore, efficiently aggregating bounding box results is also non-trivial.
Previous methods adapt the procedure via the instance-level integration to obtain image-level scores, followed by model-level averages~\cite{choi2021gmm,gal2017mcdropout,beluch2018ensemble} which cannot be applied to box-level detection. 
Or otherwise, aggregating multiple sets of box results would incur pair-wise label assignment, due to the fact that we have to traverse every prediction from all other members to construct a committee for each instance~\cite{roy2018qbc}.

Orthogonal to the model-end diversity in principle, we instead propose to introduce invariant transformations on the input-end. The posterior disagreement is thus estimated amongst multiple stochastic views of the input, which can be considered as committee members.
Drawing variations from data augmentation instead of model ensemble greatly alleviates the burden of training. 
To achieve complexity reduction for result assignment, inspired by the \textit{consensus} formulation~\cite{mccallum1998qbcconsensus,settles2012active}, we keep an exponential moving average (EMA) $\mathbb E(\theta')$ of the detector $\mathbb M(\theta)$ as a \textit{chairman} to generate box \textit{references}:
\begin{equation}
\label{eq:ema}
    \theta'_{tr} = \alpha \theta'_{tr-1} + (1-\alpha)\theta_{tr},
\end{equation}
where $tr$ indicates the training step within one cycle. 
As shown in Fig.~\ref{fig:framework}, the chairman model generates more reliable predictions~\cite{laine2017EMA} $\mathbb E(x)$ with regard to the input $x$. Meanwhile, the detector bears more diversity and produces competing hypotheses $\{b_m^j\}$ for a batch of $M$ stochastic augmentations $\mathcal{A}_m(x)$. Based on the chairman predictions as a reference, measuring disagreement between it and all other member hypotheses can effectively reduce the assignment complexity.
Note that those augmentations are fed into the network as batches and run in parallel in practice, instead of being forwarded in multiple passes.
Next, we detail our disagreement quantification for classification and localization.

\textbf{Disagreement on classification.} 
In estimating the potential value of a box to the classification branch, we prioritize controversial regions in the input space. Specifically, given the box candidates $\{b^i\}$ predicted by the chairman, 
member boxes $\{b_m^j\}$ are assigned to each reference box in $\{ b^i\}$ using, though not limited to, the detector-defined assignment strategy, such as the max-IoU assigner. 

Given a matched pair of boxes $\{ b^i, b^j_m\}$, we measure the classification disagreement based on the cross entropy between the one-hot chairman prediction $\textbf{q}^i$ and the  posterior predictive member distribution $\textbf{q}^j$:
\begin{equation}
\label{eq:ce}
    d_c^{ij}=-\mathbb{E}_{\scalebox{0.8}{$\textbf{q}^i$}}[\log \textbf{q}^j].
\end{equation}
And the disagreement about box $b^i$ is aggregated among $M$ committee members:
\begin{equation}
\label{eq:cls_uncertainty}
    d_c^i=\frac{1}{M}\sum_m^M\left(\frac{1}{k_{mi}}\sum_j^{k_{mi}} d_c^{ij}\right),
\end{equation}
where $k_{mi}$ denotes the number of positively matched member predictions in the $m$-th stochastic view.
A larger value indicates higher disagreement amongst the input-end committee over a box candidate. It shows that the current model cannot consistently make invariant label predictions under varying degrees of image perturbations, and thus it should be queried for human annotations.

\textbf{Disagreement on localization.} While it is straightforward to adopt the prediction distribution as the confidence indicator, $d_c$ can only reflect the committee disagreement on classification.
Considering the multi-tasking nature of detection, we are motivated to measure the controversy over localization.

Inspired by \cite{kao2018localizationaware,xu2021softteacher}, with multiple stochastic perturbations applied on the input, we estimate the variation of their box regression results. 
The intuition behind it is that, if the predicted position is seriously interfered due to randomness, the judgment of the current model on the target concept might not be trustworthy, and thus should be aided by human annotations. The reverse applies when the predictions remain stable despite input variations.

Specifically, with the same chairman-member label assigner used for the classification counterpart, a box reference $b^i$ is matched by multiple candidates $\{b^j_m\}$ generated by $M$ members. We apply respective inverse transformations on those boxes, which are aligned as $\{\mathcal{A}^{-1}_m(b^j_m)\}$ and fed into the localization branch of the chairman model $\mathbb E^{reg}$. Then the disagreement over the location of $b^i$ is measured based on the chairman re-calibrated boxes:
\begin{equation}
\label{eq:loc_stability}
    d_r^i = \frac{1}{4}\sum_k^4 \hat 
                 \sigma_k(\{{\mathbb E^{reg}}(\mathcal{A}^{-1}_m( b^j_m))\}).
\end{equation}
In doing so, the localization task is decomposed into four regression tasks based on coordinates. $\hat \sigma_k$ represents the standard deviation of the $k$-th coordinate, which is normalized by the average of box height and width.

Overall, for the box-level detection task, our scoring function is formulated as follows:
\begin{equation}
\label{eq:scoring}
    d^i=d_c^i \times d_r^i,
\end{equation}
based on which we rank all reference boxes for unlabeled regions, and provide labels for top-ranking boxes if they meet certain IoU-based criterion with interested targets during the annotation procedure.

Measuring controversy in both classification and localization exploits human annotations at the bounding box level. 
Built upon the scoring function, our voting committee is constructed with input-end stochasticity to avoid duplicated training, and the reference formulation further reduces assignment complexity in box-level active acquisition.
With controversial regions of the input space being efficiently identified and annotated, the generalization error minimization is gradually achieved in subsequent cycles. 

\subsection{Sparse- and Mixed-Supervision Training}
\label{sec:training}
Ever since the first active sampling, sparsely-labeled images are incorporated into the queried pool, where annotated targets provide additional information, and meantime unlabeled ones bring the noise.
More severely, our setting prioritizes challenging targets, which we empirically found to be small-sized, distant or occluded, whereas salient and dominant objects are more likely to be left unlabeled. 
As a result, the label absence of confident objects provides incorrect supervision signals, and proposals associated with them are mistakenly classified as hard negatives. If not properly handled, the sparse annotation problem would have a detrimental effect on the detection performance (See Sec.~\ref{sec:ablation}).

Despite the significant label absence, as described in Sec.~\ref{sec:active}, the silver lining is that human annotations have been provided for targets that the previous detector fails to interpret, leaving the easier ones to be concerned about.
We find the pseudo-label generation \textit{complementary} to it, where targets with confident model predictions are kept for self-training, while challenging targets with uncertain predictions are filtered out.
With both active sparse training and pseudo-label generation, we can reduce noise incurred by missing labels, as well as alleviate the error accumulation of pseudo signals.
To exploit labeled, sparsely labeled and optionally unlabeled images, we adopt the SOTA pseudo-label generation scheme inspired by \cite{liu2021unbiased,xu2021softteacher,tarvainen2017mean}.

\textbf{Supervised loss for labeled images.}
Fully labeled images $\{x^i_l\}$ from $\mathit{L}_t$ are fed into the detector and learned in a supervised way:
\begin{equation}
    \mathcal{L}_l = \frac{1}{N_l}\sum_i
    \mathcal{L}_{cls}(x_l^i, y_l^i) + 
    \mathcal{L}_{loc}(x_l^i, t_l^i),
\end{equation}
where $N_l$ is the number of fully labeled images, $y$ represents ground truth class labels and $t$ denotes corresponding box locations. $\mathcal{L}_{cls}$ and $\mathcal{L}_{loc}$ represent loss functions used by the detector for classification and localization respectively.

As described in Eq.~\ref{eq:ema}, we keep a temporal smoothed version of the detector, which is also denoted as a \textit{teacher} model. Here we refer to it as \textit{chairman} following Sec.~\ref{sec:active} for consistency.

\textbf{Pseudo-label generation.}
The data batch is appended with randomly sampled sparse or unlabeled images if available. 
The weakly augmented input image $x$ is processed by the chairman to generate pseudo-label candidates $\{b^i\}$ , while the strongly augmented version $\mathcal{A}(x)$ is fed into the detector to improve data diversity.

In accordance with our acquisition strategy, pseudo-labels for classification and localization are filtered based on different criteria to ensure precision. 
Specifically, we apply confidence thresholding with a high threshold $\lambda_c$ to obtain reliable boxes $\{\hat b_c^i\}$ for  classification. 
With regard to localization, similar as in Eq.~\ref{eq:loc_stability}, we apply positional perturbations $\mathcal{a}(b^i)$ on pseudo-labels for the chairman model to refine. Candidates with predictive fluctuations lower than  a threshold $\lambda_r$ are kept to supervise the regression head, which is denoted as $\{\hat b_r^i\}$.

\begin{figure*}[t]
  \centering
    \includegraphics[width=0.92\textwidth, trim={0 0.25cm 0 0.25cm}, clip]{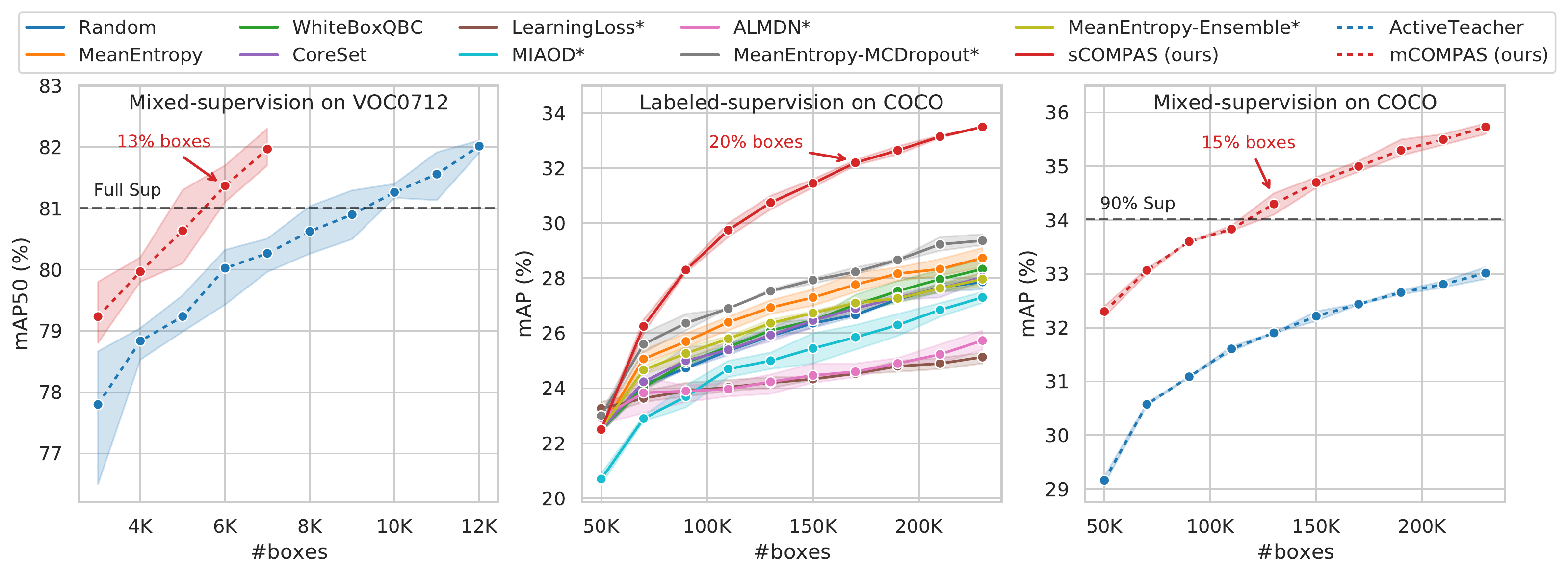}
   \caption{Box-level comparative results on (Left) VOC-semi, (Middle) COCO-sup and (Right) COCO-semi. Solid lines are performed with labeled-supervision, whereas dashed lines indicate training with unlabeled images.}
   \label{fig:mainres}
\end{figure*}
\textbf{Pseudo-active synergy for sparse images.}
Although the confidence thresholding is known to accumulate false negative errors due to the low recall of pseudo-labels, it is less likely to happen in our active sparse training setting. Because annotations of the most challenging targets have been provided. 
For a sparsely labeled image $x_s$, the pseudo-active synergy is exploited as follows:
\begin{equation}
    \mathit{G}(y_s,\hat y_{sc}) = y_s \cup \{\hat y_{sc}^i \mid \mathit{IoU}(\hat b_{sc}^i, b_s^j) \le \lambda_g, \forall
    b_s^j \in b_s\},
\end{equation}
where we supplement sparse ground truth labels $y_s$ with pseudo-labels $\hat y_{sc}$ whose corresponding boxes $\hat b_{sc}$ have less than $\lambda_g$ jaccard overlap with the ground truth ones. 
And the same de-duplication process applies to the localization branch, which results in $\mathit{G}(t_s,\hat t_{sr})$.
The supervision quality  for sparse images is thus enhanced after the completion:
\begin{equation}
\begin{aligned}
\label{eq:sparse_det}
    \mathcal{L}_s = \frac{1}{N_s}\sum_i^{N_s} 
    & \mathcal{L}_{cls}(\mathcal{A}(x_s^i), \mathit{G}(y_s^i,\hat y_{sc}^i)) + \\
    & \mathcal{L}_{loc}(\mathcal{A}(x_s^i), \mathit{G}(t_s^i,\hat t_{sr}^i)),
\end{aligned}
\end{equation}
where $N_s$ is the number of sparsely labeled images.

\textbf{Mixed-supervision with unlabeled images.}
In the pool-based active learning scenario, unlabeled images are also available during training, which can be utilized to boost performance~\cite{elezi2022NotAllLabels,mi2022ActiveTeacherSemiSupervised,yuan2021miaod}.
Without any human annotation available, the loss function is formulated as follows:
\begin{equation}
\label{eq:unlabel_det}
\begin{aligned}
    \mathcal{L}_u = \frac{1}{N_u}\sum_i^{N_u} (
    \mathcal{L}_{cls}(\mathcal{A}(x_u^i), \hat y_{uc}^i) +
    \mathcal{L}_{loc}(\mathcal{A}(x_u^i), \hat b_{ur}^i)),
\end{aligned}
\end{equation}
in which ${N_u}$ denotes the number of unlabeled images, and $\hat y_{uc}$ and $\hat b_{ur}$ are pseudo-labels and boxes for the two subtasks after thresholding respectively.

\textbf{Overall training objectives.}
In the labeled-only setting, our objective function is formulated as $\mathcal{L}_l + \frac{N_s}{N_l}\mathcal{L}_s$, where we use the sample ratio $\frac{N_s}{N_l}$ to control the contributions of the sparse data flow. 
Likewise, the objective function for the mixed-supervision setting is $\mathcal{L}_l + \frac{N_s}{N_l}\mathcal{L}_s + \frac{N_u}{N_l}\mathcal{L}_u$.

In grouping hard annotations and easy pseudo-labels together, ComPAS strategy leverages both human brainpower and machine intelligence. It frees object detectors from image-level exhaustive annotations and greatly reduces labor costs.

\tocless\section{Experiments}
\subsection{General Setup}
\textbf{Datasets.} We study previous and the proposed methods under the box-level evaluation setting on 1) PASCAL VOC0712~\cite{everingham2010voc} dataset, of which the trainval split contains 16,551 images with 40K boxes from 20 classes, and we validate on VOC07 test split; 2) Microsoft COCO~\cite{lin2014microsoft} dataset, which includes 118K images with about 860K boxes for 80 classes on the train2017 split, and 5K images for validation.

\textbf{Baselines and Evaluation.} 
Depending on the holistic involvement of unlabeled images during training, existing active learning strategies are divided into \textbf{labeled-supervised} methods (Random, MeanEntropy, WhiteBoxQBC~\cite{roy2018qbc}, CoreSet~\cite{sener2018coreset}, LearningLoss~\cite{yoo2019learningloss}, MIAOD\footnote{MIAOD~\cite{yuan2021miaod} samples an unlabeled pool of the same size as the labeled pool, and thus is excluded from holistic mixed-supervision comparison.}~\cite{yuan2021miaod}, ALMDN~\cite{choi2021gmm}, MCDropout~\cite{gal2017mcdropout}, Ensemble~\cite{beluch2018ensemble} and our supervised-ComPAS (denoted as sComPAS)) and \textbf{mixed-supervised} ones (ActiveTeacher~\cite{mi2022ActiveTeacherSemiSupervised} and our mixed-ComPAS (mComPAS)). 
Under different supervision and datasets, we refer to our experimental settings as \textbf{VOC-sup, VOC-semi, COCO-sup} and \textbf{COCO-semi}.

On VOC0712 dataset, to initialize the labeled pool, we randomly sample images for exhaustive annotation until the budget of 3K boxes is reached, and append 1K boxes per cycle. 
On COCO, images of 50K boxes are randomly sampled and annotated at first, and 20K boxes are labeled per cycle based on respective query strategies. 
We conduct all experiments in the main paper for 10 cycles unless the fully supervised (FS) performance has been reached.
We report mean average precision @0.5 (mAP50) for VOC0712 and  @0.5:0.95 (mAP) for COCO.  The mean and standard deviation of results for three independent runs are reported.

\textbf{Implementation.} 
Our detector implementation and training configurations are based on Faster R-CNN~\cite{ren2015faster} with ResNet-50~\cite{he2016resnet} backbone under the mmdetection~\cite{chen2019mmdetection} codebase. For a fair comparison, we re-implement MeanEntropy, WhiteBoxQBC~\cite{roy2018qbc}, CoreSet~\cite{sener2018coreset}, LearningLoss~\cite{yoo2019learningloss}, MIAOD~\cite{yuan2021miaod}, ALMDN~\cite{choi2021gmm}, MCDropout~\cite{gal2017mcdropout} and Ensemble~\cite{beluch2018ensemble} based on their respective public code (if available) or paper descriptions\footnote{We verify that the performance of our re-implementations with sufficient regularization and augmentation can surpass their reported results.}. Details of their implementations can be found in the supplementary material.

In each independent run, the exact same data split is used for all methods, among which methods with specialized architectures have different initial performances (marked with *).
During the training of each cycle, we train 12500 or 88000 iterations with a batch-size of 16 for VOC0712 or COCO datasets respectively to be consistent with the fully supervised setting.
Unless otherwise stated, SGD optimizer is adopted with learning rates set as 0.01 or 0.02 for VOC0712 or COCO, which is decayed by 10 at 8/12$\times$ and 11/12$\times$ total iterations. 
On VOC0712, we train from scratch in each cycle, whereas for COCO we fine-tune from the previous checkpoint for 0.3$\times$ iterations with 0.1$\times$ learning rate.
In terms of the semi-supervision training of competitors and our mixed-supervision setting, we double the training iterations following the common practice, and leave the rest unchanged.
For the proposed method, thresholds $\lambda_c, \lambda_r, \lambda_g$ are set as $0.9, 0.02$ and $0.4$ respectively, which are not specially tuned. We adopt $M$=10 committee members.
Diverse augmentations (\eg flip, color distortion) are applied for all methods to make full use of available data.

All experiments on VOC0712 were conducted on NVIDIA RTX 3090, and those of COCO were performed on Tesla V100.

\begin{figure}[tp]
    \centering
    \hspace*{-0.5cm}\includegraphics[width=.36\textwidth, trim={0 0.2cm 0 0.2cm}, clip]{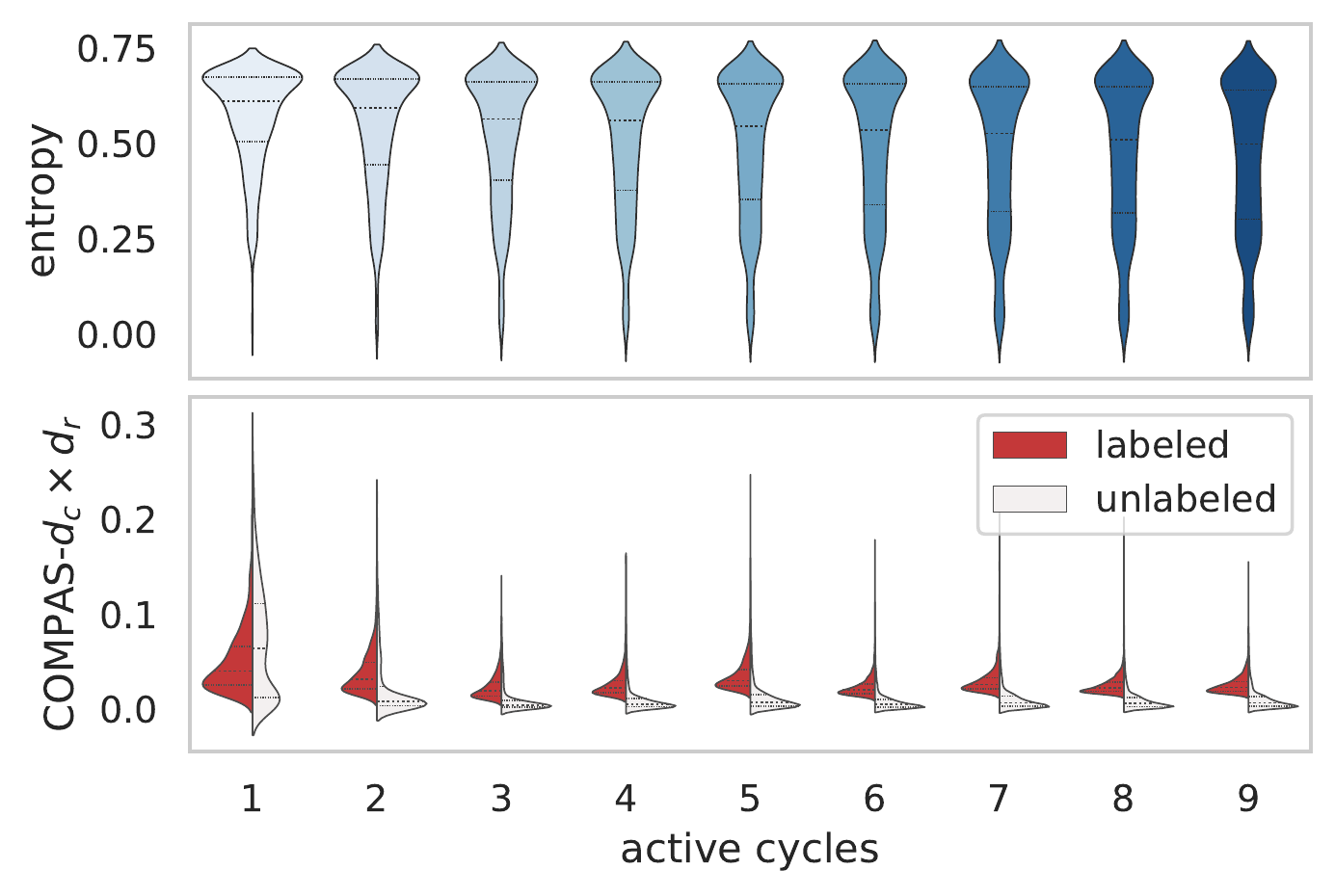}
   \caption{Violin plots showing the box-level score distributions of newly acquired images in each active learning cycle calculated by (Top) image-level MeanEntropy and (Bottom) box-level ComPAS. }
   \label{fig:violin}
\end{figure}

\subsection{Main Results}
\textbf{Image-level vs. box-level evaluation}. 
The comparison between the image-level and box-level evaluation settings under VOC-sup is shown in Fig.~\ref{fig:image-level}. Although most of the SOTAs and our hacking method BoxCnt work well under the image-level evaluation, their scoring functions obscurely prioritize crowded images, or their highly ranked targets are severely interfered by invaluable counterparts from the same images. Thus, when evaluated under the box-level criterion, resources wasted on the latter ones emerge, and some previous conclusions are no longer tenable.

\textbf{Performance comparison}. 
Results under VOC-sup and COCO-sup are presented in Fig.~\ref{fig:image-level}R and Fig.~\ref{fig:mainres}M respectively.
As can be seen, within the same box-based budgets, the proposed method outperforms baselines and SOTAs at each active learning cycle by a large margin. Under the labeled-supervision setting for VOC, we obtain 100\% supervised performance with only 9K ground truth boxes, which efficiently saves approximately 81\% label expenditure.
The superiority of our method is also clearly demonstrated on COCO, where sComPAS can exploit rich knowledge from both human annotations and model intelligence. It consistently beats the second-best model-end ensemble-based method by a large margin, and outperforms it by 4.3\% mAP in the last cycle in a robust and efficient manner.

In leveraging the unlabeled pool, as shown in Fig.~\ref{fig:mainres}LR, we first notice that 
the proposed active learning strategy retains its overall supremacy: surpassing the 100\% supervised performance requires less than 13\% boxes for mComPAS on VOC0712, and on the COCO dataset it only requires 15\% boxes to achieve 90\% fully supervised  capability. 
In comparison, 
ActiveTeacher~\cite{mi2022ActiveTeacherSemiSupervised} adopts an advanced semi-supervised model~\cite{sohn2020simple} for pseudo-label generation, but its acquisition function is solely based on predictive class distributions, which cannot exploit the pseudo-active synergy or well capture sample informativeness.

Under four benchmark settings, the consistent improvements of ComPAS over active learning cycles demonstrate the effectiveness of the input-end committee in identifying informative targets to benefit the detector. Built upon it, our superior results over competitors further show that the proposed pipeline can maximize return over investment by the pseudo-active synergy on the box-level.

\begin{figure}[tp]
    \centering
    \hspace*{-0.5cm}\includegraphics[width=.45\textwidth, trim={0 0.2cm 0 0.2cm}, clip]{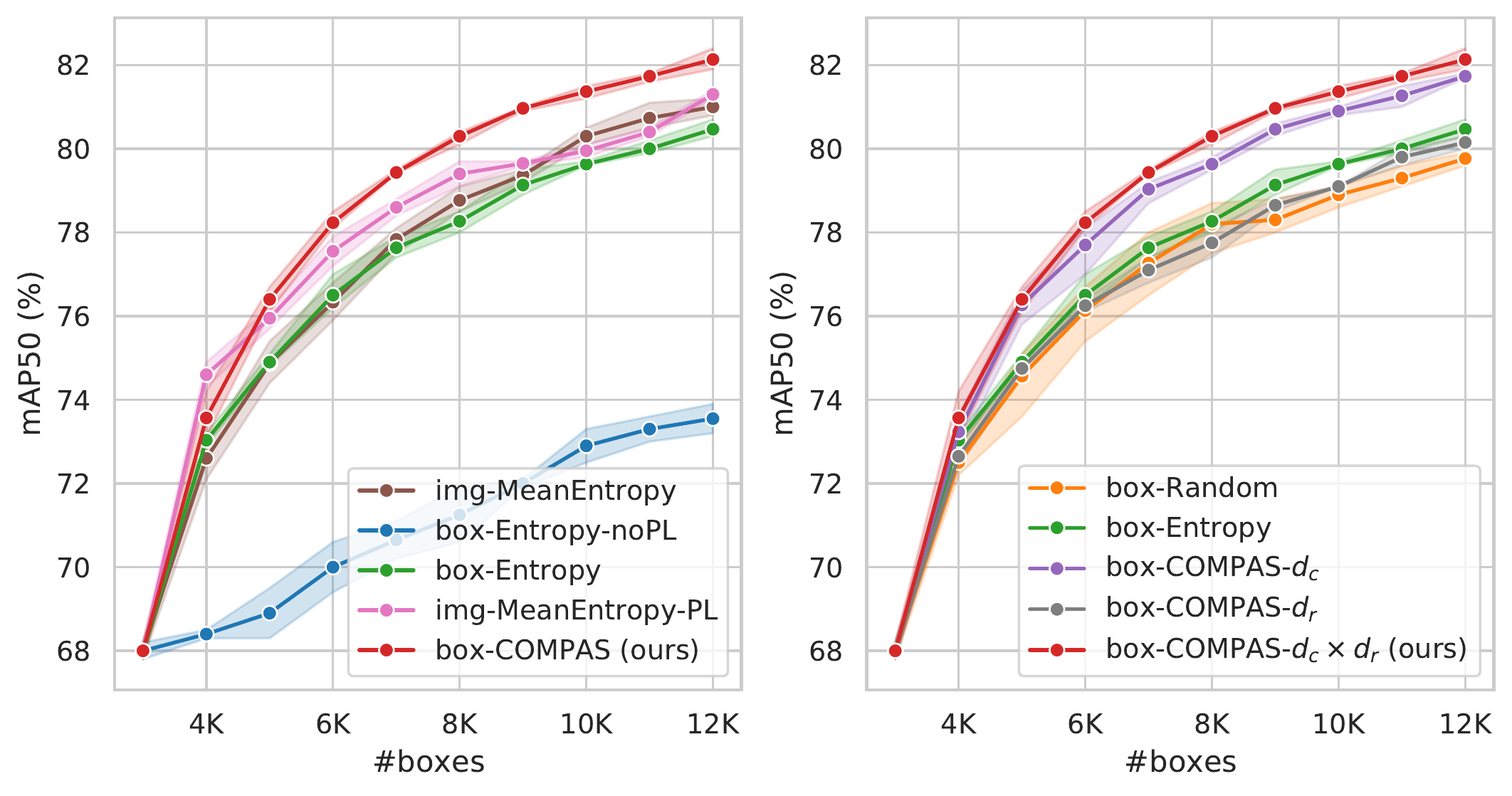}
   \caption{Analysis on (Left) the extension from image-level annotation to the box-level and (Right) alternatives to the box-level active acquisition strategy on VOC0712.}
   \label{fig:ablation}
\end{figure}

\subsection{Quantitative Analysis}
\label{sec:ablation}

\textbf{Redundant annotations in the image-level annotation}.
We take MeanEntropy sampling, the best image-level single-model method, to demonstrate the redundancy problem. In Fig.~\ref{fig:violin}, MeanEntropy shows a long-tail phenomenon in the score distributions of newly acquired images, which gets even more acute in later cycles. It indicates that the scoring functions of image-level methods are interfered by less informative targets. Passively annotating them along with highly-ranked ones results in redundancy. 
In contrast, our box-level method actively annotates valuable targets and leaves the rest unlabeled, maximizing the return over investment. As the iteration proceeds, the divergence between distributions of labeled and unlabeled boxes consistently increases, demonstrating the improvement of model capability and acquisition reliability. 
 
Next, to show the extension of annotation protocol from image-level to the box-level, in Fig.~\ref{fig:ablation}L, we take MeanEntropy sampling as a baseline, and apply the box-level annotation protocol, ComPAS model design choices and our scoring function step-by-step under the VOC-sup setting.

\textbf{Impact of box-level sparse annotation}. 
Image-level exhaustive annotation (\textit{img-MeanEntropy}), despite the label redundancy, guarantees the stable training of detection models.
When the annotation is disentangled into the box-level, without specific handling (\textit{box-Entropy-noPL}), the missing label problem has a detrimental effect on the model due to the incorrect supervision signal. 
To alleviate the problem, we introduce pseudo-label generation for sparse images (\textit{box-Entropy}), where sparse labels for challenging targets are supplemented with confident model predictions via an IoU-based grouping strategy, which rectifies supervision signals and significantly boosts performance.
However, we notice that it is outperformed by the image-level counterpart in later cycles, which indicates that the box-level annotation poses a greater challenge to the budget allocation, under which entropy-based sampling is not an optimal informativeness estimation solution. Thus, we propose \textit{box-ComPAS}, which is analyzed later in this section.

\begin{figure}[tbp]
  \centering
    \hspace*{-0.1cm}\includegraphics[width=.48\textwidth]{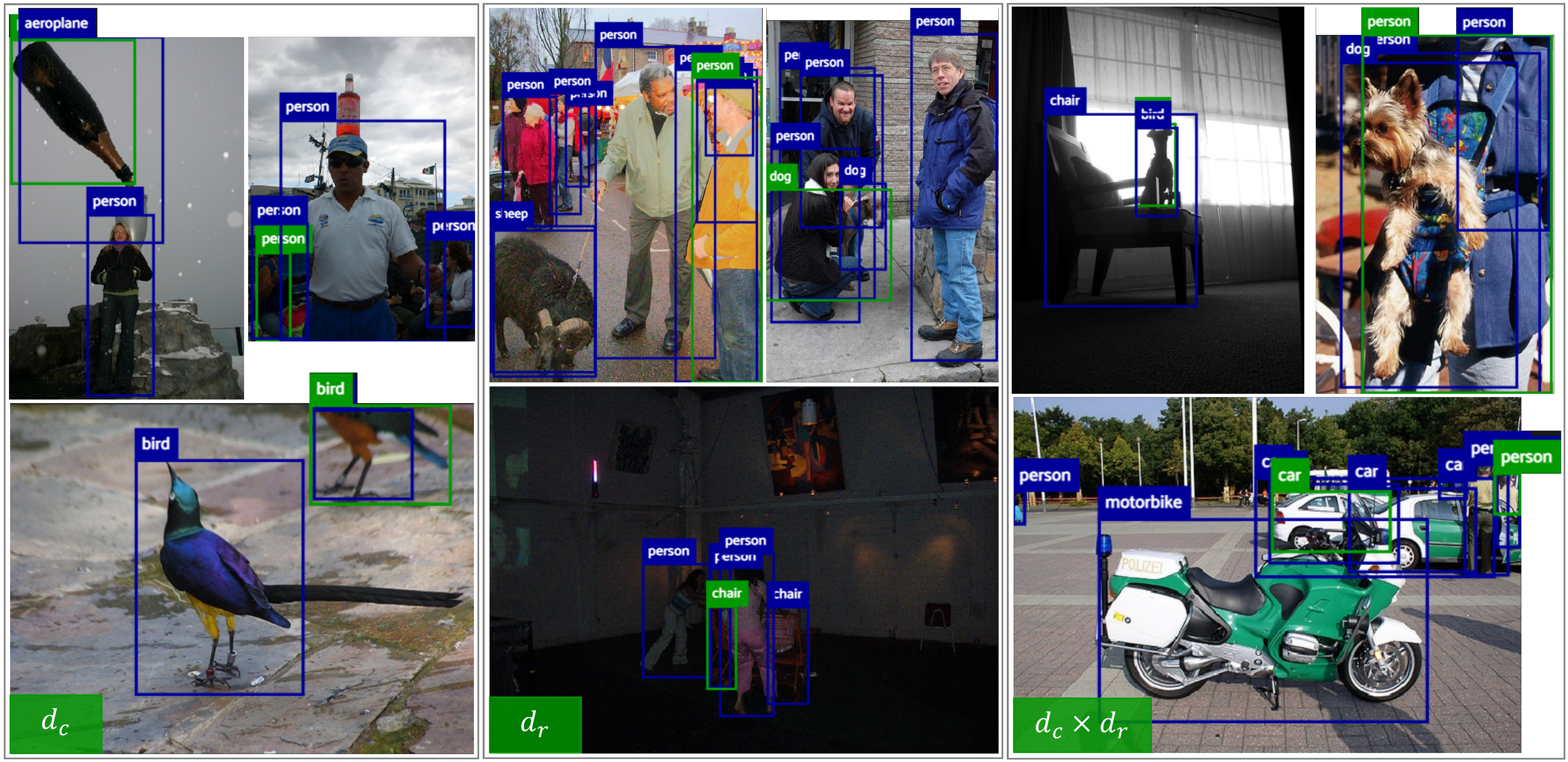}
  \caption{Qualitative results of targets top-ranked by our scoring functions (in green) and complementary pseudo-labels (in blue).} 
   \label{fig:visuals}
\end{figure}

\textbf{ComPAS model design}.
The pseudo-label generation scheme designed for sparse annotations of box-level methods can also be used to boost the performance for fully supervised training. To present the effect of it, we simply apply it on all labeled images for image-level MeanEntropy (\textit{img-MeanEntropy}). We first observe that pseudo-labeling is especially effective in the low data regime, but the performance increment  is limited in later cycles as the knowledge grows. We also note that our method retains superiority although only sparse images are fed for pseudo-labeling, which demonstrates that the effectiveness of ComPAS is attributed to informative box selection, while pseudo-labeling is mainly used to compensate for acquired knowledge.

\textbf{Box-level scoring function}.
Under the box-level annotation protocol, we experiment with alternatives to our  scoring function during the active acquisition stage, which includes Random, Entropy, our classification disagreement estimation $d_c$ in Eq.~\ref{eq:cls_uncertainty} alone, our localization disagreement estimation $d_r$ in Eq.~\ref{eq:loc_stability} alone, and the proposed classification-localization hybrid metric $d_c \times d_r$ presented in Eq.~\ref{eq:scoring}. All alternatives are performed with the input-end committee ensemble same as ours.
As the results in Fig.~\ref{fig:ablation}R suggest, baseline methods, such as entropy-based sampling that performs well for image-level annotation, are not optimal box-level informativeness indicators. In contrast, $d_c$ estimates the cross entropy between the consensus and member prediction distributions, which well captures the classification informativeness. 
But built upon $d_c$ and $d_r$, our hybrid metric further incorporates disagreement estimation about the localization subtask, benefiting both detection heads from human annotations. 
It shows that our acquisition function reflects the challenge of boxes being correctly and robustly detected given the current level of knowledge, so that highly ranked boxes can play a complementary role with pseudo-labels in the subsequent training cycles.

\subsection{Qualitative Analysis.}
The complementarity between actively queried targets (in green) and pseudo-labels (in blue) are visualized in Fig.~\ref{fig:visuals}. We present top-ranked boxes scored by our classification metric $d_c$, localization metric $d_r$ as well as the hybrid metric $d_c\times d_r$ respectively, and give the chairman-generated pseudo-labels from the same learning cycle.
We empirically find that actively queried targets are more likely to be small, occluded or deviant, where the model fails to guarantee invariant predictions under strong input variations.
In contrast, targets left by our scoring function tend to be salient and ubiquitous, whose online pseudo-labels usually get better and better in the next cycles and play a complementary role. 
More visualizations are shown in the supplementary.

\tocless\section{Conclusion}
In this paper, we reveal the pitfalls of image-level evaluation for active detection and propose a realistic and fair box-level evaluation criterion. We then advocate efficient box-level annotation, under which we formulate a novel active detection pipeline, namely Complementary Pseudo Active Strategy (ComPAS) to exploit both human annotations and machine intelligence. It evaluates box informativeness based on the disagreement amongst a near-free input-end committee for both classification and localization to effectively query challenging targets. Meantime, the detector model addresses the sparse training problem by pseudo-label generation for well-learned targets. Under both labeled-only and mixed-supervision settings on VOC0712 and COCO datasets, ComPAS outperforms competitors by a large margin in a unified codebase. 

\tocless\section{Acknowledgment}
This work was partly supported by National Key R\&D Program of China (No. 2022ZD0119400), National Natural Science Foundation of China (Nos. 61925107, 62271281, U1936202), Zhejiang Provincial Natural Science Foundation of China under Grant (No. LDT23F01013F01), China Postdoctoral Science Foundation (BX2021161) and Tsinghua-OPPO JCFDT.

\newpage
{\small
\bibliographystyle{ieee_fullname}
\bibliography{egbib}
}

\clearpage
\setcounter{section}{0}
\setcounter{figure}{0}
\setcounter{table}{0}

\twocolumn[{%
\renewcommand\twocolumn[1][]{#1}%
\vspace{-3em}
\begin{center}
    \centering
    \Large{\textbf{Box-Level Active Detection -- Supplementary Material}}
\end{center}%
  \begin{figure}[H]
  \setlength{\linewidth}{\textwidth}
  \setlength{\hsize}{\textwidth}
  \centering
  \includegraphics[width=1\textwidth]{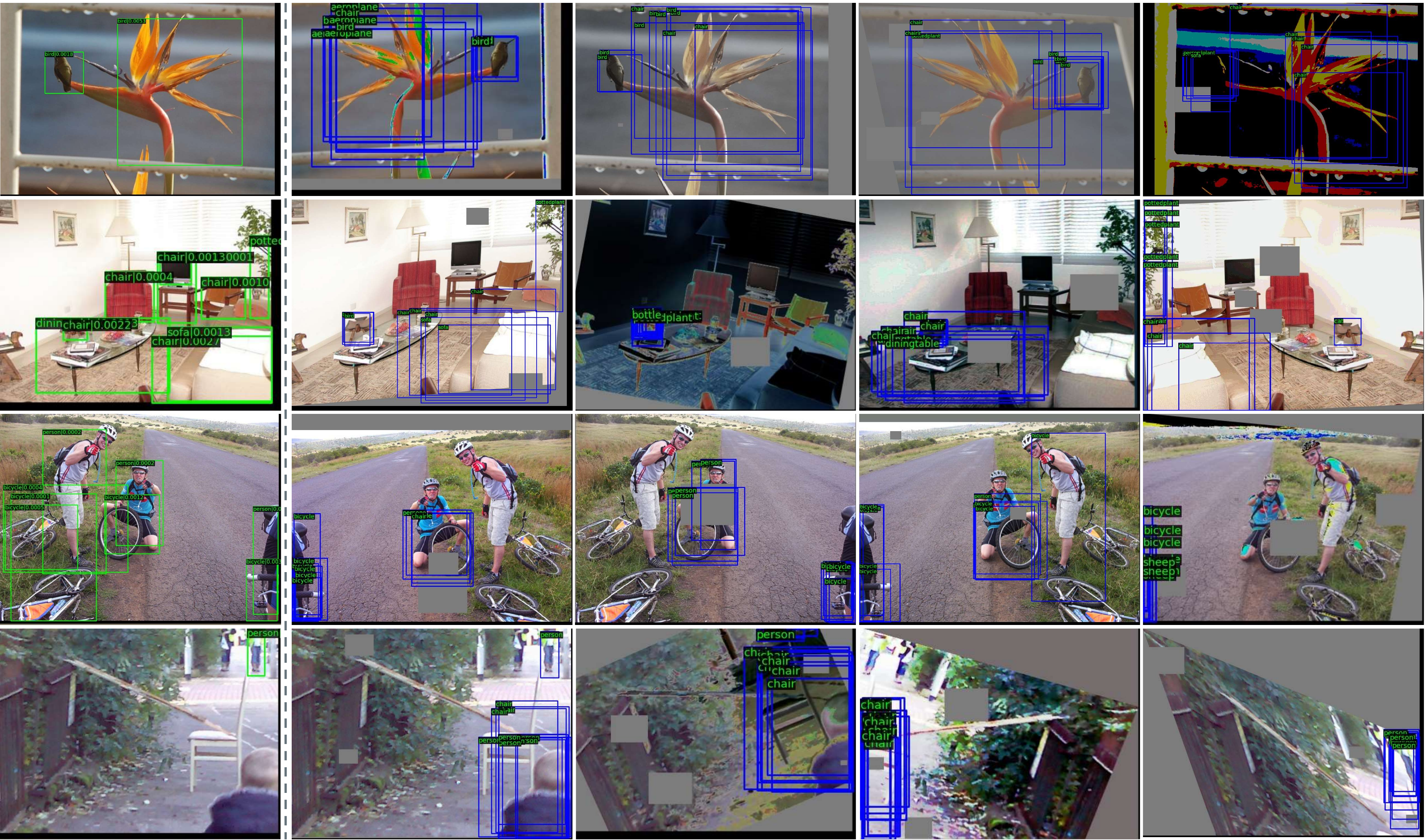}
  \caption{Reference generated by the chairman model (1st column with boxes in green and box-level acquisition scores), and corresponding member predictions (2nd-5th columns with boxes in blue). Experiments are performed at the first active learning cycle under the VOC-sup setting. We only show top-ranking member predictions for ease of visualization.}
  \label{fig:ref_voc}
  \end{figure}
\vspace{1em}
}]

{\hypersetup{linkcolor=blue}
\tableofcontents}

\begin{figure*}[ht]
\centering
\includegraphics[width=1\textwidth]{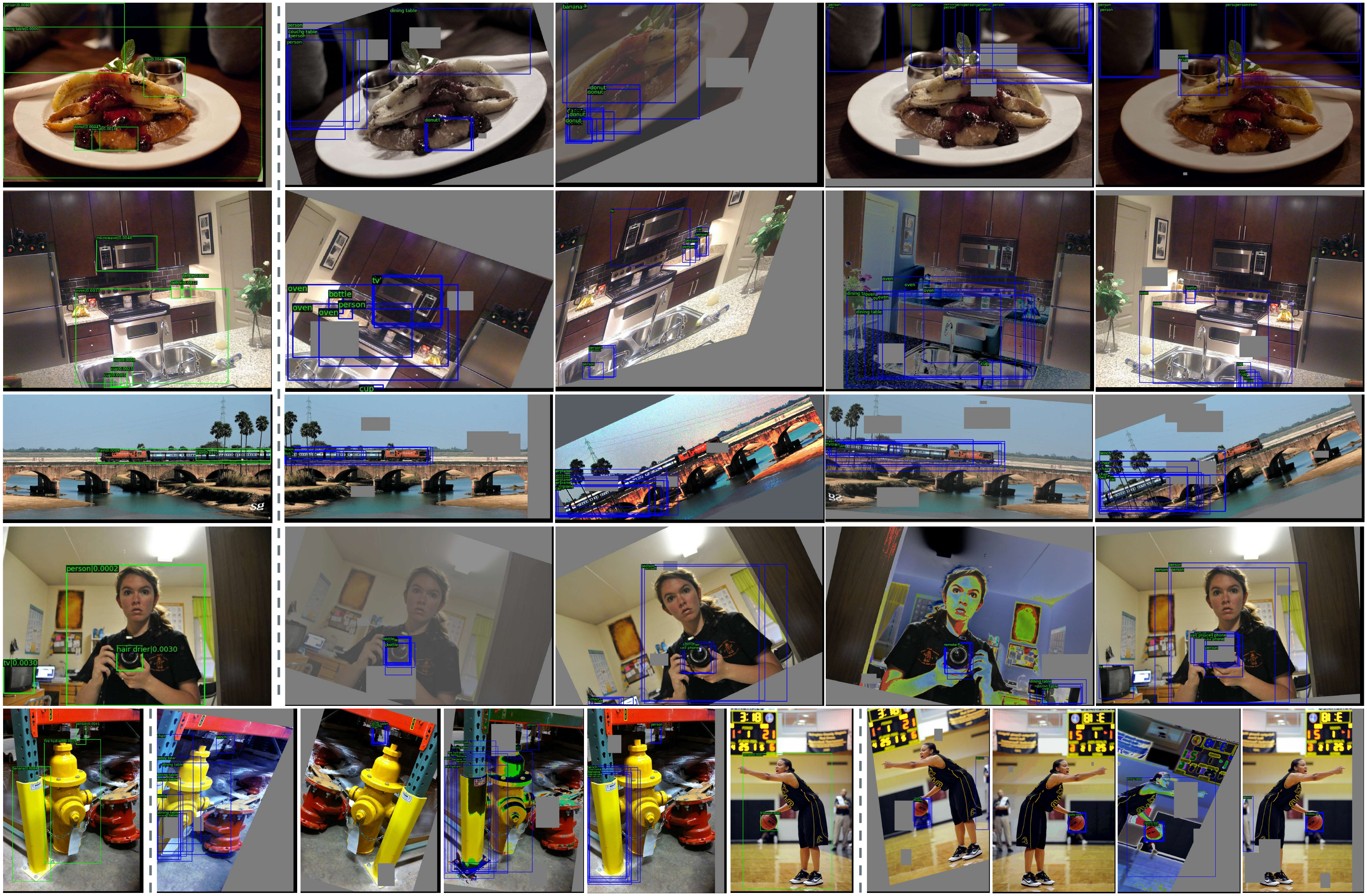}
\caption{Reference generated by the chairman model (1st column with boxes in green and box-level acquisition scores), and corresponding member predictions (2nd-5th columns with boxes in blue). Experiments are performed at the first active learning cycle under the COCO-sup setting. We only show top-ranking member predictions for ease of visualization.}
\label{fig:ref_coco}
\end{figure*}
  
\begin{table*}[htp]
\centering
\resizebox{1\linewidth}{!}{ 
\begin{tabular}{c|ccccccccccc}
\toprule
sComPAS Ablations & 0 & 1 & 2 & 3 & 4 & 5 & 6 & 7 & 8 & 9 & Inference Time (s) \\
\midrule
Members2Members & 68.00$\pm$0.20 & 73.15$\pm$0.41 & 75.85$\pm$1.26 & 78.00$\pm$0.20 & 78.60$\pm$0.20 & 79.50$\pm$0.34 & 80.00$\pm$0.06 & 80.55$\pm$0.27 & 81.10$\pm$0.20 & 81.20$\pm$0.20 & 0.6739 \\
DetectorRef2Members & 68.00$\pm$0.20 & 73.40$\pm$0.34 & 76.35$\pm$0.55 & 77.70$\pm$0.34 & 78.75$\pm$0.41 & 79.50$\pm$0.20 & 79.75$\pm$0.13 & 79.95$\pm$0.27 & 80.10$\pm$0.14 & 80.70$\pm$0.28 & 0.1656 \\
ChairmanRef2Members (ours) & 68.00$\pm$0.20 & \textbf{73.57$\pm$0.57} & \textbf{76.40$\pm$0.30} & \textbf{78.23$\pm$0.23} & \textbf{79.43$\pm$0.06} & \textbf{80.30$\pm$0.17} & \textbf{80.97$\pm$0.06} & \textbf{81.37$\pm$0.15} & \textbf{81.73$\pm$0.12} & \textbf{82.13$\pm$0.25} & 0.1667 \\
\bottomrule
\end{tabular}}
\caption{Ablation study on the existence and the source of the acquisition reference under the VOC-sup setting. The accuracy (\%) and time consumption are averaged over 3 runs. Inference Time in seconds denotes the average forward time per image during the acquisition stage.}
\label{tab:ref}
\end{table*}

\section{Efficacy of the Acquisition Reference}
In this section, we validate the use of acquisition reference in terms of both accuracy and efficiency. The comparison is conducted under the VOC-sup setting on the same server with 4 NVIDIA RTX 3090. We set the committee size as 10 in accordance with our main results.

\textbf{The existence of the reference}. 
Without the reference, the \textit{Members2Members} variant constructs a committee for each instance by traversing every prediction from all other members, as done in WhiteBoxQBC~\citesupp{suproy2018qbc}.
In contrast, with a reference incorporated, the assignment procedure can be reduced to comparing between it and other member hypotheses.
As shown in Tab.~\ref{tab:ref}, if facilitated by the reference, such as in our \textit{ChairmanRef2Members}, the inference for acquisition takes about 0.1667 seconds per image in practice. It is approximately $4\times$ faster than the Members2Members ablation, which indicates that the reference is essential when scaling up to larger datasets.
Besides, since the augmentations used to construct the committee is diverse and relatively strong, basing the informativeness estimation solely on member outputs is susceptible to noise and randomness. As the detection accuracy shows, with a robust and reliable reference, the quality of active sampling consistently improves over cycles.

\textbf{The source of the reference}. 
To obtain reliable references as query candidates, the original image is fed into the chairman model, whose predictions are adopted as reference in our proposed ComPAS design. For comparison with the chairman, we experiment with the \textit{DetectorRef2Members} variant, where the reference is obtained from the detector itself instead of its temporal ensemble. As the results in Tab.~\ref{tab:ref} show, with almost equal computational cost, the quality of chairman-generated references beat the non-EMA alternative by a large margin.

\textbf{Visual inspection of our reference}. In Fig.~\ref{fig:ref_voc} and Fig.~\ref{fig:ref_coco}, we visualize top-ranked reference hypotheses according to our disagreement scores, as well as member predictions assigned to them. The experiment is performed at the 1st cycle under the VOC-sup and COCO-sup settings respectively. For ease of visualization, we trim member predictions based on their scores, \ie, contributions to the acquisition score of the reference box, so that only top-ranking box predictions are shown. 
We observe that the chairman model can recognize well-learned targets and locate some salient objects as reference in a stable manner, whereas those perturbed members can bring considerable variations and randomness. Member predictions challenge the judgments of the chairman or supplement with potential candidates, so that controversial regions of the input space can be found. 
For example, in Fig.~\ref{fig:ref_voc}, the bird in the first row obtains a consensus, whereas the branch next to it is mistakenly recognized by the chairman as a bird with spread wings, which is disputed by the committee. Similar observations can also be made from the camera held by the girl shown in Fig.~\ref{fig:ref_coco}.
Besides, the proposed metric also prioritizes targets that are challenging to localize. Take the 3rd row in Fig.~\ref{fig:ref_coco} for example: the committee has a disagreement over the train body, which leads to a higher acquisition score.
The qualitative results further indicate that the proposed disagreement quantification under strong variations well identifies the input space where the current model neglects. Once actively annotated, they can effectively provide informativeness and guarantee consistent improvements in later model updates. 

\begin{table}[!t]\small
    \centering
    \resizebox{0.7\linewidth}{!}{%
    \begin{tabular}{c|ccc}
        \hline
        \multirow{2}{*}{Cycle} & \multicolumn{3}{c}{M} \\ \cline{2-4} 
        & 1 & 4 & 10 \\ \hline
        0 & 68.00$\pm$0.20 & 68.00$\pm$0.20 & 68.00$\pm$0.20 \\
        1 & 72.97$\pm$0.35 & \textbf{73.83$\pm$0.47} & 73.57$\pm$0.57 \\
        2 & 76.37$\pm$0.21 & \textbf{76.83$\pm$0.32} & 76.40$\pm$0.30 \\
        3 & 78.10$\pm$0.46 & 78.17$\pm$0.38 & \textbf{78.23$\pm$0.23} \\
        4 & 78.93$\pm$0.31 & 79.20$\pm$0.10 & \textbf{79.43$\pm$0.06} \\
        5 & 79.67$\pm$0.31 & 79.90$\pm$0.35 & \textbf{80.30$\pm$0.17} \\
        6 & 80.33$\pm$0.12 & 80.50$\pm$0.35 & \textbf{80.97$\pm$0.06} \\
        7 & 80.87$\pm$0.47 & 80.83$\pm$0.25 & \textbf{81.37$\pm$0.15} \\
        8 & 81.37$\pm$0.12 & 81.33$\pm$0.23 & \textbf{81.73$\pm$0.12} \\
        9 & 81.70$\pm$0.26 & 81.83$\pm$0.15 & \textbf{82.13$\pm$0.25} \\
        \hline
    \end{tabular}%
    }
    \caption{Sensitivity to the input-end committee size under the VOC-sup setting.}
    \label{table:size}
\end{table}

\section{Sensitivity to the Input-end Committee Size}
As there is no consensus on the appropriate committee size to use~\citesupp{supsettles2012active}, we experiment under the VOC-sup setting with a varying number of members $M$. As shown in Tab.~\ref{table:size}, although one member prediction can work well under the proposed pipeline, providing more data variations on the input-end helps the model identify more informative and representative samples and provides robustness, which guarantees further improvements and stability in a cheap but effective way.

\section{Comparison with Ensemble-based Methods}

We further compare the proposed ComPAS with well-performed ensemble-based methods, including Ensemble~\citesupp{supbeluch2018ensemble} and MCDropout~\citesupp{supgal2017mcdropout}. Following \citesupp{supchoi2021gmm}, those multi-model methods are implemented based on MeanEntropy, which is also the best single-model baseline in our experiments.
The detection accuracy has been presented in Fig.~1, Fig.~3 and Tab.~1 of the main paper, and the numerical results of figures are reported in Tab.~\ref{tab:voc} and Tab.~\ref{tab:coco} respectively. In Tab.~\ref{tab:efficiency}, we detail the size of the ensemble/committee, 
the required training time and the inference time per image for informativeness estimation. The experiments are conducted under the VOC-sup setting on the same server with 4 NVIDIA RTX 3090. 

As can be seen, even when there is only one member, \ie $M=1$, our method retains its overall supremacy in both effectiveness and efficiency.
Built upon it, we provide flexibility in the committee size to suit the needs of different end applications. With more members incorporated, better detection performance can be further pursued via more input variations. Since the committee construction only happens during the acquisition stage, and image perturbations can be processed in one feed-forward pass in practice, the extra costs incurred are marginal in contrast to ensemble~\citesupp{supbeluch2018ensemble} and MCDropout~\citesupp{supgal2017mcdropout}.

\begin{table}[tp]
\centering
\resizebox{1\linewidth}{!}{ 
\begin{tabular}{c|ccc}
\toprule
Methods & M & \#Trainable Parameters (M) & Inference Time (s)\\
\midrule
MeanEntropy-Ensemble    & 3  & 123.51 & 0.0310 \\
MeanEntropy-MCDropout   & 25 & 41.17  & 0.2400 \\
\midrule
\multirow{4}{*}{sComPAS (ours)}  & 1  & 41.17  & 0.0277 \\
            & 4  & 41.17  & 0.0741 \\
            & 10 & 41.17  & 0.1667 \\
            & 15 & 41.17  & 0.2582 \\
\bottomrule
\end{tabular}}
\caption[]{Comparison with the ensemble-based methods under the VOC-sup setting. $M$ represents the size of the ensemble/committee following the implementation of \citesupp{supchoi2021gmm,supbeluch2018ensemble}. The number of trainable parameters are reported in millions. Inference Time in seconds denotes the average forward time per image during the acquisition stage.}\label{tab:efficiency}
\end{table}

\begin{figure}[tp]
\centering
\includegraphics[width=0.35\textwidth]{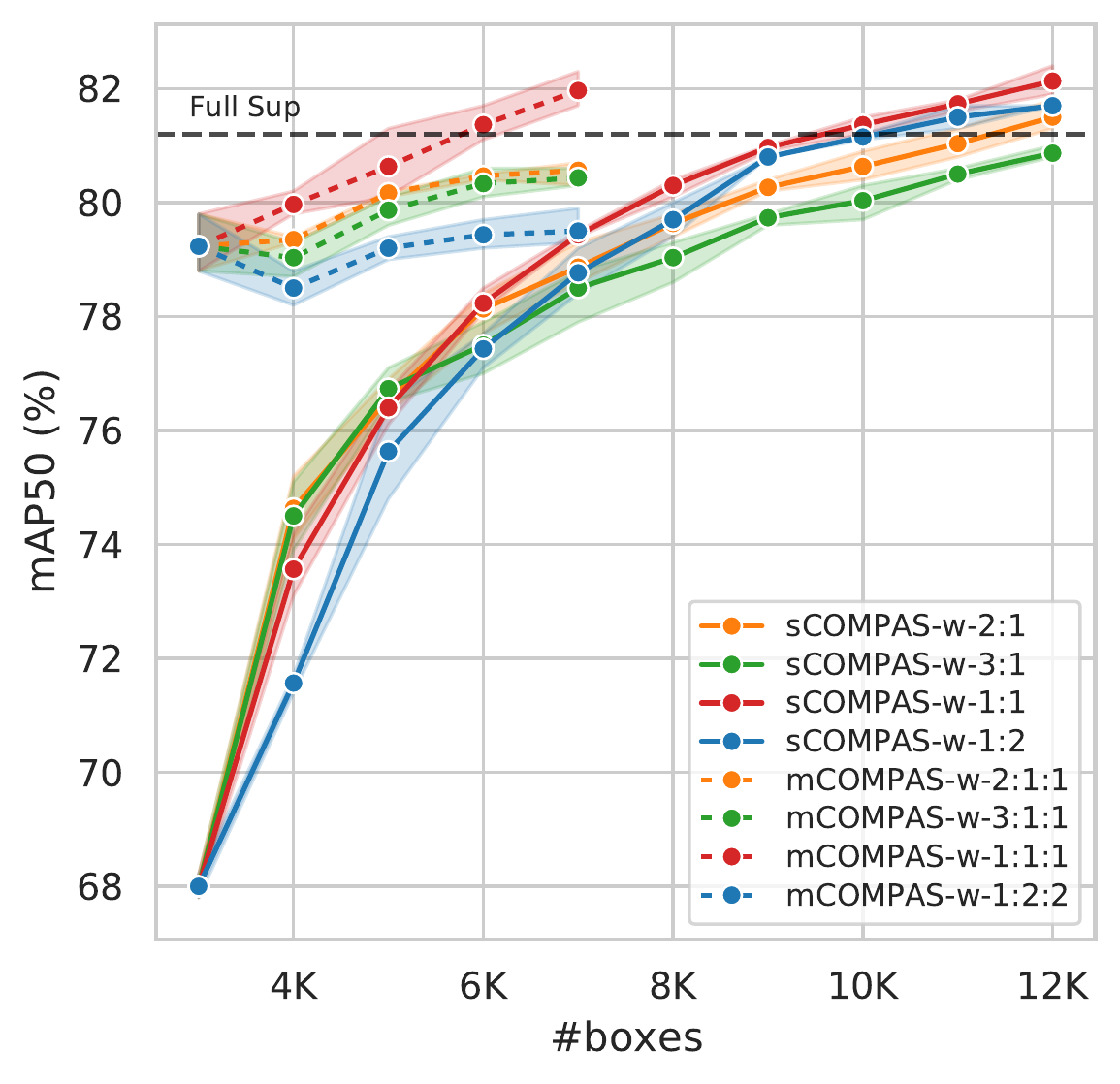}
\caption{Analysis of the loss weight ratio among different levels of supervision (full-labeled: partial-labeled(: unlabeled)) under the VOC-sup and VOC-semi settings.}
\label{fig:weights}
\end{figure}

\section{Analysis of Overall Loss Weights}
Our overall objective functions for both training settings give the same relevance to different levels of supervision. To analyze the importance of the fully labeled subset versus others, we specifically finetune the weight of them under both VOC-sup and VOC-semi settings.

Results in Fig.~\ref{fig:weights} show that fully labeled images are more informative when they dominate the data pool, but increasing their relevance cannot guarantee consistent improvement as the distribution changes along learning cycles. In contrast, re-weighting losses by the sample ratio keeps the method simple but effective.

\section{Performance under Different Active Learning Settings}
As active learning is known to be sensitive to settings~\citesupp{supmunjal2022RobustReproducibleActive}, in this section, we validate our observations under different scenarios.

\begin{figure}[tp]
\centering
\hspace*{-0.5cm}\includegraphics[width=.35\textwidth, trim={0 0.2cm 0 0.2cm}, clip]{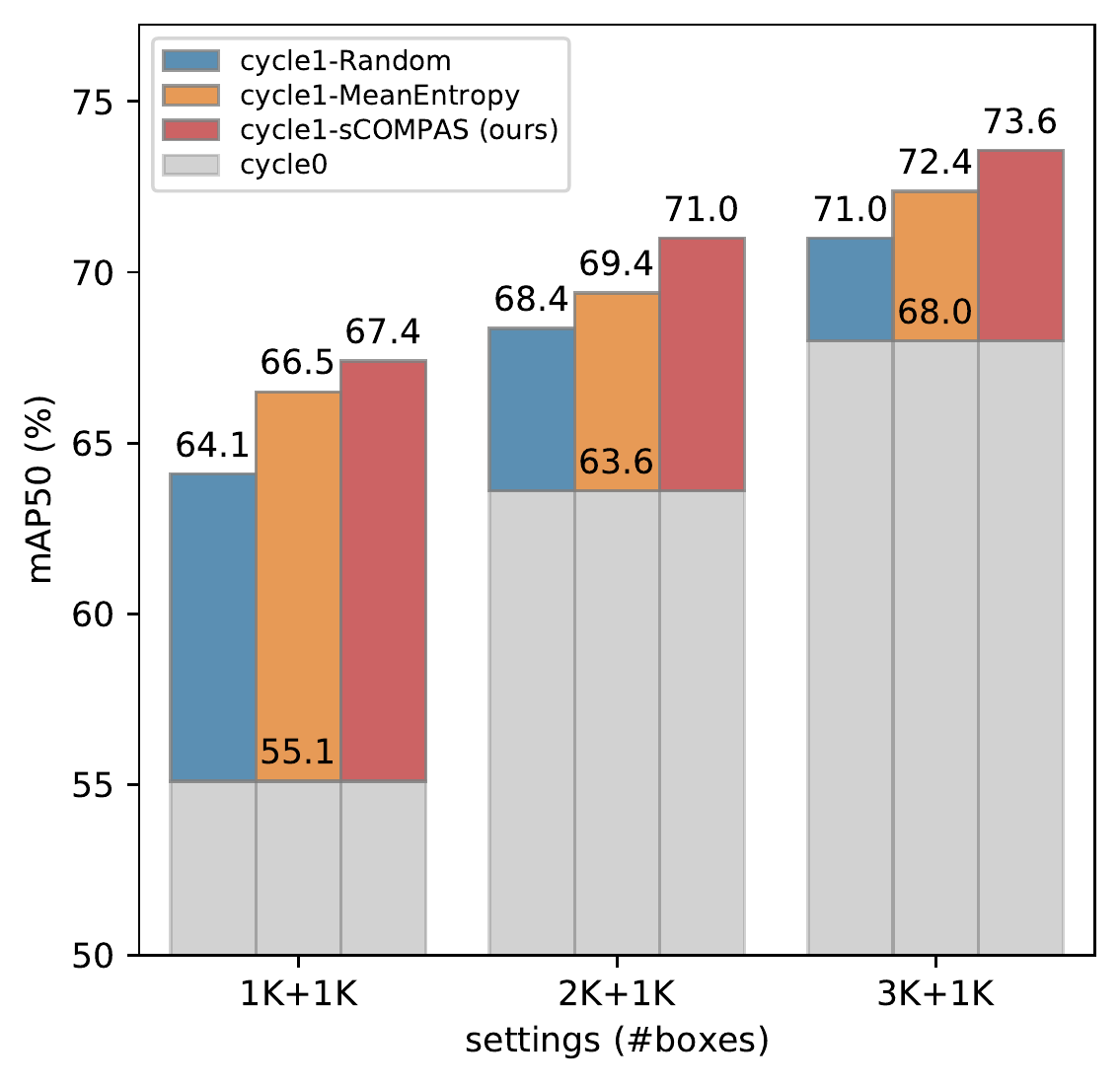}
\caption{Weaker initialization (cycle0) and the started state (cycle1) under lower-data regimes under the VOC-sup setting.}
\label{fig:lowdata}
\end{figure}

\textbf{Performance with weaker starting points}. 
In Fig.~\ref{fig:lowdata}, we start the detector under lower-data regimes without overfitting. Results of the initialization with 3K, 2K and 1K boxes and of the corresponding started states demonstrate that our active sampling strategy is robust to weaker initialization and outperforms competitors. 

\begin{figure}[tp]
\centering
\hspace*{-0.5cm}\includegraphics[width=.35\textwidth, trim={0 0.2cm 0 0.2cm}, clip]{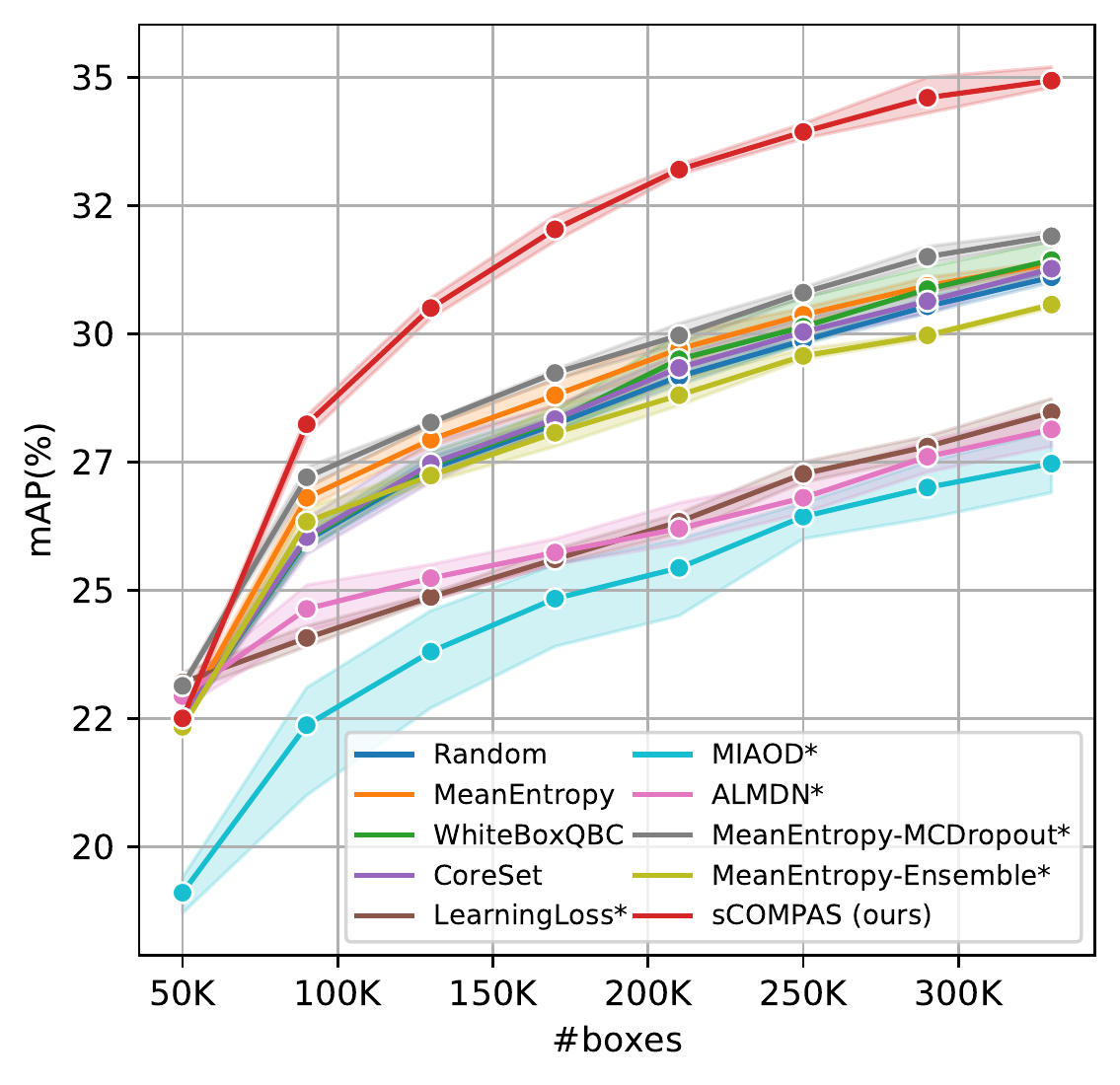}
\caption{Box-level comparative results on COCO-sup with a 40K per-cycle budget.}
\label{fig:coco54}
\end{figure}

\textbf{Performance with a larger acquisition batch size}.
We conduct experiments on the COCO dataset under the COCO-sup setting with a larger acquisition batch size. The iterations are initialized with 50K boxes, the same as the experiments shown in Fig.~3M of the main paper. In each active learning cycle, we append 40K boxes based on respective query and annotation strategies. 
Results of three independent runs are plotted in Fig.~\ref{fig:coco54} and numerically detailed in Tab.~\ref{tab:coco54} respectively, which show that the proposed method is superior regardless of the acquisition batch size.

\begin{figure*}[p]
\centering
\includegraphics[width=1\textwidth]{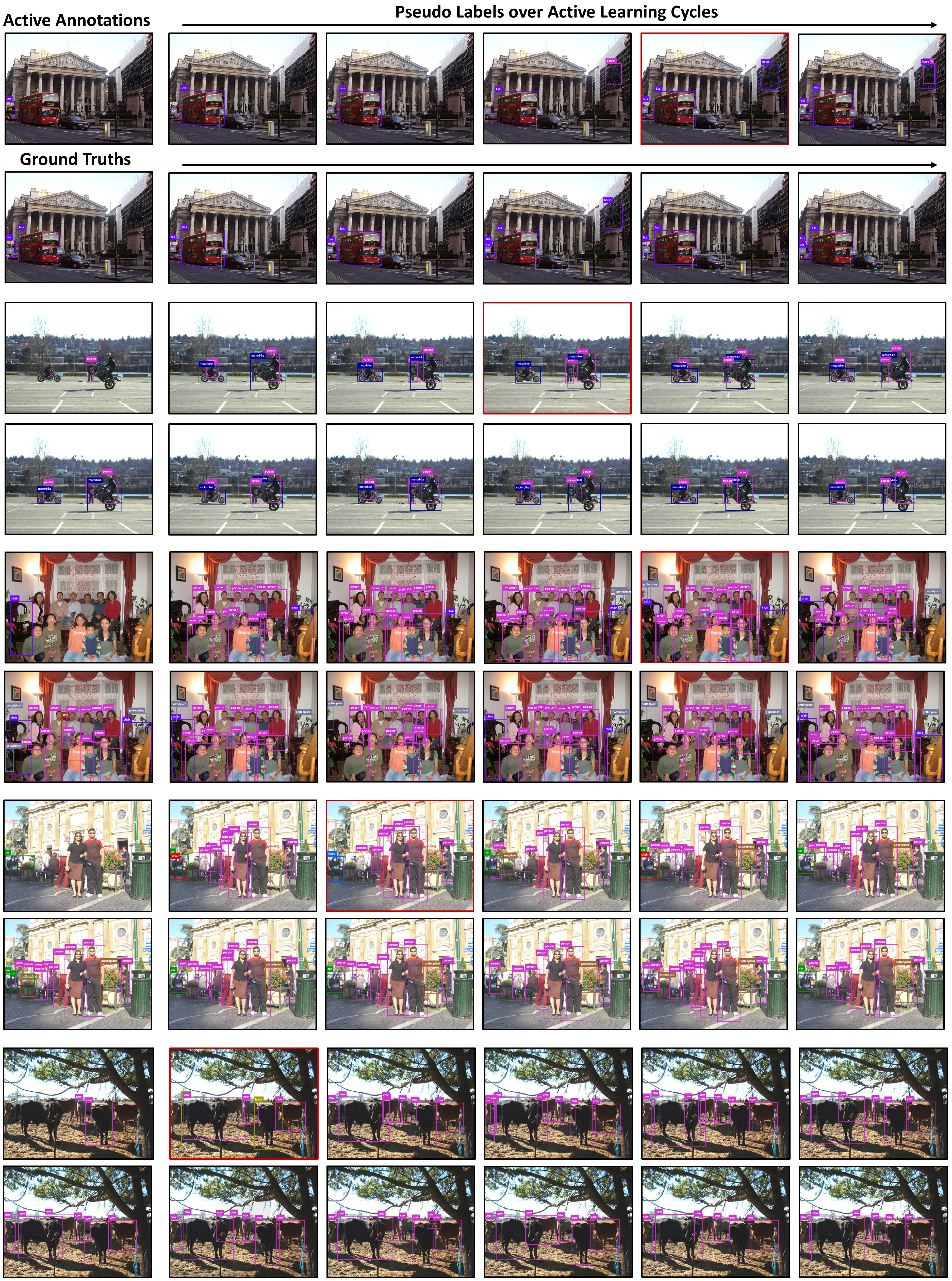}
\caption{Iteration of pseudo labels across active learning cycles accompanied by actively annotated targets. For each set of images, we highlight the image in red frame if the active annotation happens in that step.}
\label{fig:PseudoEvolve}
\end{figure*}

\begin{table*}[t]
\centering
\resizebox{1\linewidth}{!}{ 
\begin{tabular}{c|cccccccccc}
\toprule
\multirow{2}{*}{Methods} & 0 & 1 & 2 & 3 & 4 & 5 & 6 & 7 & 8 & 9 \\
\cmidrule{2-11}
 & \multicolumn{10}{c}{VOC-sup}\\
\midrule
Random & 68.00$\pm$0.20 & 71.10$\pm$0.56 & 73.03$\pm$0.60 & 74.23$\pm$0.38 & 75.93$\pm$0.31 & 77.07$\pm$0.23 & 77.77$\pm$0.21 & 78.60$\pm$0.10 & 79.23$\pm$0.15 & 79.80$\pm$0.10 \\
MeanEntropy & 68.00$\pm$0.20 & 72.60$\pm$0.44 & 74.87$\pm$0.50 & 76.33$\pm$0.45 & 77.83$\pm$0.31 & 78.77$\pm$0.31 & 79.37$\pm$0.15 & 80.30$\pm$0.20 & 80.73$\pm$0.32 & 81.00$\pm$0.20 \\
WhiteBoxQBC & 68.00$\pm$0.20 & 71.37$\pm$0.85 & 73.23$\pm$0.31 & 74.33$\pm$0.47 & 75.30$\pm$0.78 & 76.73$\pm$0.74 & 77.17$\pm$0.68 & 78.13$\pm$0.45 & 78.67$\pm$0.65 & 79.20$\pm$0.26 \\
CoreSet & 68.00$\pm$0.20 & 71.20$\pm$0.50 & 73.23$\pm$0.32 & 75.00$\pm$0.53 & 76.03$\pm$0.35 & 77.17$\pm$0.47 & 77.97$\pm$0.40 & 78.93$\pm$0.42 & 79.47$\pm$0.45 & 79.87$\pm$0.31 \\
LearningLoss* & 66.70$\pm$0.89 & 68.10$\pm$1.25 & 69.77$\pm$0.87 & 70.43$\pm$0.81 & 71.33$\pm$0.87 & 72.47$\pm$0.59 & 72.97$\pm$0.45 & 73.57$\pm$0.51 & 74.57$\pm$0.40 & 74.73$\pm$0.91 \\
ALMDN* & 67.50$\pm$0.71 & 70.00$\pm$0.28 & 70.85$\pm$0.07 & 72.05$\pm$0.64 & 73.10$\pm$0.71 & 73.65$\pm$0.92 & 74.35$\pm$0.92 & 75.25$\pm$0.92 & 75.45$\pm$1.06 & 76.35$\pm$0.78 \\
MeanEntropy-MCDropout* & \textbf{68.87$\pm$0.59} & 73.17$\pm$0.25 & 75.27$\pm$0.15 & 76.90$\pm$0.20 & 78.07$\pm$0.35 & 79.00$\pm$0.10 & 79.70$\pm$0.00 & 80.40$\pm$0.35 & 80.77$\pm$0.15 & 81.40$\pm$0.35 \\
MeanEntropy-Ensemble* & 67.87$\pm$0.23 & 72.07$\pm$0.25 & 74.70$\pm$0.20 & 75.80$\pm$0.44 & 77.30$\pm$0.10 & 78.43$\pm$0.31 & 79.30$\pm$0.17 & 79.80$\pm$0.17 & 80.70$\pm$0.26 & 81.13$\pm$0.15 \\
BoxCnt & 68.00$\pm$0.20 & 69.87$\pm$0.32 & 71.20$\pm$0.56 & 72.27$\pm$0.45 & 73.20$\pm$0.56 & 74.00$\pm$0.17 & 74.83$\pm$0.32 & 75.53$\pm$0.49 & 76.50$\pm$0.46 & 77.17$\pm$0.32 \\
sCOMPAS (ours) & 68.00$\pm$0.20 & \textbf{73.57$\pm$0.57} & \textbf{76.40$\pm$0.30} & \textbf{78.23$\pm$0.23} & \textbf{79.43$\pm$0.06} & \textbf{80.30$\pm$0.17} & \textbf{80.97$\pm$0.06} & \textbf{81.37$\pm$0.15} & \textbf{81.73$\pm$0.12} & \textbf{82.13$\pm$0.25} \\
\midrule
 & \multicolumn{10}{c}{VOC-semi}\\
\midrule
ActiveTeacher & 77.80$\pm$1.15 & {78.84$\pm$0.27} & {79.24$\pm$0.31} & {80.02$\pm$0.52} & {80.27$\pm$0.28} & {80.62$\pm$0.39} & {80.90$\pm$0.40} & {81.26$\pm$0.12} & {81.56$\pm$0.40} & {82.01$\pm$0.10} \\
mCOMPAS (ours) & \textbf{79.23$\pm$0.51} & \textbf{79.97$\pm$0.21} & \textbf{80.63$\pm$0.61} & \textbf{81.37$\pm$0.31} & \textbf{81.97$\pm$0.31}  & - &  - & -  & -  & - \\
\bottomrule
\end{tabular}}
\caption{MAP50 and standard deviation (\%) under VOC-sup and VOC-semi settings, with 3K boxes for initialization and 1K boxes for each active learning cycle. Results reported are averaged over 3 independent runs. We stop the learning of mCOMPAS after it far exceeds the fully supervised performance (81\% mAP50). The best result of each cycle is highlighted in bold. }
\label{tab:voc}
\end{table*}

\section{Pseudo-Active Synergy Visualization over Active Learning Cycles}
In Fig.~\ref{fig:PseudoEvolve}, We present the iteration of pseudo-labels predicted by the chairman model accompanied by active human annotations across the learning cycles. 
Images are highlighted in red frames if the active annotation happens in those steps.
We find that the proposed acquisition function prioritizes challenging targets, such as small, occluded (\eg cars in the 1st and 4th images) or deviant (\eg the cow in the last images mistakenly recognized as a horse) ones. Meantime, most unlabeled targets can be covered by pseudo-label generation, which gets better in both classification and localization across active learning cycles.
Through iterative knowledge gain and self-supervision, the synergy between them is effectively exploited.

\section{Implementation Details}
In addition to the general settings of our unified codebase introduced in the main paper, we detail the re-implementation of compared methods in this section.

\textbf{FullSup}. For reference, we also report the fully supervised (FS) performance, denoted by FullSup, given the same model, augmentations and runtime settings as in AL experiments. Thus $r$\% Sup refers to $r$\% of the FS performance instead of a data split.

\textbf{MeanEntropy}. 
After NMS (Non-maximal suppression), we calculate the entropy of a box candidate based on the predictive probability of its most confident class. Then the uncertainty scores are averaged over the image.

\textbf{WhiteBoxQBC}~\citesupp{suproy2018qbc}. Following the Algorithm. 1 provided in the paper, we take the NMS outputs of the multiple scales of Faster R-CNN to construct the `committee', among which all classes and pair-wise bounding box predictions are traversed to estimate the image uncertainty based on predictive \textit{margin}.

\textbf{CoreSet}~\citesupp{supsener2018coreset}. We apply global average pooling on the multi-level features extracted by the Feature Pyramid Network(FPN) of Faster R-CNN, which are then concatenated as the latent space representation of an image. We implement the k-Center-Greedy algorithm to select unlabeled images during the acquisition stage.

\textbf{LearningLoss}~\citesupp{supyoo2019learningloss}. The multi-level features extracted by the Feature Pyramid Network (FPN) of Faster R-CNN are fed into the loss prediction module. The gradient from the loss prediction module is stopped at $0.8\times$ total iterations following the implementation of the paper. We finetune the weight of the learned loss and use 0.3 in consideration of convergence and better performance.

\begin{table*}[ht]
\centering
\resizebox{1\linewidth}{!}{ 
\begin{tabular}{c|cccccccccc}
\toprule
\multirow{2}{*}{Methods} & 0 & 1 & 2 & 3 & 4 & 5 & 6 & 7 & 8 & 9 \\
\cmidrule{2-11}
 & \multicolumn{10}{c}{COCO-sup}\\
\midrule
Random & 22.50$\pm$0.10 & 24.13$\pm$0.12 & 24.73$\pm$0.06 & 25.37$\pm$0.21 & 25.90$\pm$0.17 & 26.37$\pm$0.15 & 26.67$\pm$0.06 & 27.23$\pm$0.06 & 27.67$\pm$0.15 & 27.87$\pm$0.23 \\
MeanEntropy & 22.50$\pm$0.10 & 25.07$\pm$0.29 & 25.70$\pm$0.26 & 26.40$\pm$0.26 & 26.93$\pm$0.25 & 27.30$\pm$0.30 & 27.77$\pm$0.38 & 28.17$\pm$0.25 & 28.33$\pm$0.32 & 28.73$\pm$0.35 \\
WhiteBoxQBC & 22.50$\pm$0.10 & 24.03$\pm$0.15 & 24.93$\pm$0.15 & 25.50$\pm$0.10 & 26.10$\pm$0.26 & 26.47$\pm$0.06 & 27.03$\pm$0.32 & 27.53$\pm$0.40 & 27.97$\pm$0.42 & 28.33$\pm$0.47 \\
CoreSet & 22.50$\pm$0.10 & 24.23$\pm$0.25 & 25.00$\pm$0.44 & 25.40$\pm$0.17 & 25.93$\pm$0.25 & 26.47$\pm$0.31 & 26.90$\pm$0.17 & 27.30$\pm$0.10 & 27.63$\pm$0.29 & 28.03$\pm$0.15 \\
LearningLoss* & \textbf{23.27$\pm$0.25} & 23.63$\pm$0.23 & 23.90$\pm$0.26 & 24.03$\pm$0.23 & 24.20$\pm$0.20 & 24.33$\pm$0.06 & 24.53$\pm$0.06 & 24.80$\pm$0.20 & 24.90$\pm$0.20 & 25.13$\pm$0.21 \\
MIAOD* & 20.70$\pm$0.28 & 22.90$\pm$0.14 & 23.70$\pm$0.57 & 24.70$\pm$0.42 & 25.00$\pm$0.42 & 25.45$\pm$0.78 & 25.85$\pm$0.64 & 26.30$\pm$0.57 & 26.85$\pm$0.35 & 27.30$\pm$0.28 \\
ALMDN* & 22.93$\pm$0.21 & 23.83$\pm$0.67 & 23.90$\pm$0.35 & 23.97$\pm$0.25 & 24.23$\pm$0.38 & 24.47$\pm$0.38 & 24.60$\pm$0.26 & 24.90$\pm$0.17 & 25.23$\pm$0.35 & 25.73$\pm$0.35 \\
MeanEntropy-MCDropout* & 23.00$\pm$0.17 & 25.60$\pm$0.36 & 26.37$\pm$0.31 & 26.90$\pm$0.00 & 27.53$\pm$0.06 & 27.93$\pm$0.12 & 28.23$\pm$0.15 & 28.67$\pm$0.06 & 29.23$\pm$0.25 & 29.37$\pm$0.21 \\
MeanEntropy-Ensemble* & 22.53$\pm$0.12 & 24.67$\pm$0.15 & 25.27$\pm$0.25 & 25.80$\pm$0.10 & 26.37$\pm$0.12 & 26.73$\pm$0.06 & 27.10$\pm$0.10 & 27.27$\pm$0.06 & 27.63$\pm$0.15 & 27.97$\pm$0.25 \\
sCOMPAS (ours) & 22.50$\pm$0.10 & \textbf{26.25$\pm$0.35} & \textbf{28.30$\pm$0.14} & \textbf{29.75$\pm$0.35} & \textbf{30.75$\pm$0.35} & \textbf{31.45$\pm$0.21} & \textbf{32.20$\pm$0.14} & \textbf{32.65$\pm$0.21} & \textbf{33.15$\pm$0.07} & \textbf{33.50$\pm$0.00} \\
\midrule
 & \multicolumn{10}{c}{COCO-semi}\\
\midrule
ActiveTeacher & 29.16$\pm$0.10 & 30.57$\pm$0.04 & 31.09$\pm$0.01 & 31.61$\pm$0.07 & 31.90$\pm$0.01 & 32.21$\pm$0.13 & 32.44$\pm$0.02 & 32.66$\pm$0.05 & 32.81$\pm$0.08 & 33.01$\pm$0.16 \\
mCOMPAS (ours) & \textbf{32.30$\pm$0.10} & \textbf{33.07$\pm$0.06} & \textbf{33.60$\pm$0.00} & \textbf{33.83$\pm$0.06} & \textbf{34.30$\pm$0.20} & \textbf{34.70$\pm$0.10} & \textbf{35.00$\pm$0.10} & \textbf{35.30$\pm$0.17} & \textbf{35.50$\pm$0.10} & \textbf{35.73$\pm$0.12} \\
\bottomrule
\end{tabular}}
\caption{MAP and standard deviation (\%) under COCO-sup and COCO-semi settings, with 50K boxes for initialization and 20K boxes for each active learning cycle. Results reported are averaged over 3 independent runs. The best result of each cycle is highlighted in bold. }
\label{tab:coco}
\end{table*}

\textbf{ALMDN}~\citesupp{supchoi2021gmm} Following the code published by the authors, we turn the detection heads of Faster R-CNN into four Gaussian mixture models (GMMs), and take the maximum over epistemic and aleatoric uncertainty for classification and localization as the score of an unlabeled image.

\textbf{MIAOD}~\citesupp{supyuan2021miaod} Following the code published by the authors, the second detection head and an additional multi-instance learning (MIL) head are built upon Faster R-CNN, and the disagreement between the two classification branches is used as an uncertainty indicator. The weight of MIL is finetuned as 0.1 in consideration of convergence and better performance. 
The results of MIAOD obtained from our re-implementation significantly surpass those reported by the authors on the COCO dataset.
However, under the VOC setting, consistent improvement cannot be assured over the active learning cycles after parameter search, among which the best performance we can get is no more than 72\% mAP50. Experiments with the code provided by the authors under the same budget also confirm our observation, where we only get around 70\% mAP50 in the last cycle. This might suggest that MIAOD is less applicable to the low-data regime. Thus, the results of MIAOD on the VOC dataset are not reported in the main paper.

\textbf{MCDropout}~\citesupp{supgal2017mcdropout}. Following the practice in \citesupp{supchoi2021gmm}, the adaption of MCDropout to the detection task is achieved by image-level estimation followed by averaging the results of 25 forward passes. Dropout layers with $p=0.1$ are inserted at the last two stages after each bottleneck module of the ResNet backbone.
The image-level informativeness can be estimated by different acquisition functions. Here we also follow the implementation in \citesupp{supchoi2021gmm} to use entropy, which performs the best under our experimental settings.

\textbf{Ensemble}~\citesupp{supbeluch2018ensemble}. Similarly, we establish an ensemble of three independent detectors following \citesupp{supchoi2021gmm}. The informativeness estimation and result ensemble are the same as the MCDropout implementation.

\textbf{BoxCnt}  BoxCnt is devised by us to attack the image-level estimation for active detection. This hack is achieved by naively prioritizing unlabeled images with the most number of box predictions after NMS.

\begin{table*}[ht]
\centering
\resizebox{0.9\linewidth}{!}{ 
\begin{tabular}{c|cccccccccc}
\toprule
\multirow{2}{*}{Methods} & 0 & 1 & 2 & 3 & 4 & 5 & 6 & 7  \\
\cmidrule{2-9}
 & \multicolumn{8}{c}{COCO-sup}\\
\midrule
    Random & 22.50$\pm$0.10 & 25.97$\pm$0.15 & 27.37$\pm$0.21 & 28.23$\pm$0.15 & 29.17$\pm$0.21 & 29.87$\pm$0.06 & 30.53$\pm$0.15 & 31.10$\pm$0.10 \\
    MeanEntropy & 22.50$\pm$0.10 & 26.80$\pm$0.20 & 27.93$\pm$0.23 & 28.80$\pm$0.26 & 29.70$\pm$0.30 & 30.37$\pm$0.15 & 30.93$\pm$0.15 & 31.37$\pm$0.06 \\
    WhiteBoxQBC & 22.50$\pm$0.10 & 26.00$\pm$0.20 & 27.43$\pm$0.23 & 28.30$\pm$0.20 & 29.50$\pm$0.46 & 30.13$\pm$0.49 & 30.87$\pm$0.38 & 31.43$\pm$0.32 \\
    CoreSet & 22.50$\pm$0.10 & 26.03$\pm$0.35 & 27.47$\pm$0.40 & 28.33$\pm$0.23 & 29.33$\pm$0.25 & 30.03$\pm$0.23 & 30.63$\pm$0.21 & 31.27$\pm$0.12 \\
    LearningLoss* & \textbf{23.20$\pm$0.20} & 24.07$\pm$0.21 & 24.87$\pm$0.12 & 25.60$\pm$0.17 & 26.33$\pm$0.21 & 27.27$\pm$0.21 & 27.80$\pm$0.20 & 28.47$\pm$0.40 \\
    MIAOD* & 19.10$\pm$0.36 & 22.37$\pm$1.18 & 23.80$\pm$0.98 & 24.83$\pm$0.83 & 25.43$\pm$0.81 & 26.43$\pm$0.38 & 27.00$\pm$0.56 & 27.47$\pm$0.60 \\
    ALMDN* & 22.93$\pm$0.21 & 24.63$\pm$0.45 & 25.23$\pm$0.38 & 25.73$\pm$0.25 & 26.20$\pm$0.44 & 26.80$\pm$0.36 & 27.60$\pm$0.30 & 28.13$\pm$0.35 \\
    MeanEntropy-MCDropout* & 23.13$\pm$0.06 & 27.20$\pm$0.20 & 28.27$\pm$0.06 & 29.23$\pm$0.12 & 29.97$\pm$0.21 & 30.80$\pm$0.10 & 31.50$\pm$0.20 & 31.90$\pm$0.10 \\
    MeanEntropy-Ensemble* & 22.33$\pm$0.06 & 26.33$\pm$0.23 & 27.23$\pm$0.15 & 28.07$\pm$0.46 & 28.80$\pm$0.26 & 29.57$\pm$0.12 & 29.97$\pm$0.06 & 30.57$\pm$0.06 \\
    sCOMPAS (ours) & 22.50$\pm$0.10 & \textbf{28.23$\pm$0.21} & \textbf{30.50$\pm$0.20} & \textbf{32.03$\pm$0.25} & \textbf{33.20$\pm$0.10} & \textbf{33.93$\pm$0.15} & \textbf{34.60$\pm$0.36} & \textbf{34.93$\pm$0.23} \\
\bottomrule
\end{tabular}}
\caption{MAP and standard deviation (\%) under COCO-sup and COCO-semi settings, with 50K boxes for initialization and 20K boxes for each active learning cycle. Results reported are averaged over 3 independent runs. The best result of each cycle is highlighted in bold. }
\label{tab:coco54}
\end{table*}

\section{Related Work Beyond Active Learning}
The main paper discusses widely used and most recent acquisition functions, ensemble models and evaluation methods for active learning. In addition to it, we give a more extensive survey of literature related to the proposed framework and method in this section.

\textbf{Consistency regularization}. 
Active learning estimates predictive consistency as an uncertainty indicator to sample unlabeled candidates, based on which images with the most inconsistent hypotheses are queried for human annotation. To this end, previous methods measure the \textit{disagreement} between multiple models or heads~\citesupp{supseung1992qbc,supbeluch2018ensemble,supyuan2021miaod,supchoi2021gmm,suproy2018qbc}, and the \textit{robustness} of the output after noise perturbation~\citesupp{supkao2018localizationaware} or horizontal flip~\citesupp{supelezi2022NotAllLabels} of the image.

While the consistency estimation of active learning mainly happens in the acquisition stage, quite similarly, semi-supervised learning (SSL)~\citesupp{supoliver2018realisticconsistency,supsohn2020simple,supjeong2019consistency,supxu2021softteacher} encourages consistency in the outputs of realistic image perturbations during training. The shared motivation behind them is to find a smooth manifold for the dataset so that the version space of the model is minimized~\citesupp{supsettles2012active,supbelkin2006manifold,supoliver2018realisticconsistency}.
Their goals are further aligned in \citesupp{supgao2020consistencyeccv} for active classification, where under the semi-supervised learning framework, Gao \etal sample images with inconsistent predictions that the model has difficulty self-learning via consistency regularization.
Most recent active detection methods also draw on advanced semi-supervised learning (SSL) techniques. For example, Mi \etal~\citesupp{supmi2022ActiveTeacherSemiSupervised} achieve active sampling under the framework of the unbiased teacher~\citesupp{supliu2021unbiased}, a state-of-the-art semi-supervised detection method to exploit all available data. But their informativeness estimation is based on entropy and class diversity, which is not aligned with the training objective. 
Elezi \etal~\citesupp{supelezi2022NotAllLabels} measures the Kullback-Leibler divergence between the predictive distributions of flipped images. They also provide offline pseudo-labels for unlabeled images with confident predictions, so that human-labeled, pseudo-labeled and unlabeled images are all involved in consistency-based model optimization. 

In comparison to previous studies, the proposed ComPAS method provides sufficient perturbations as the testing ground for disagreement estimation, aligns the consistency-oriented goal in model training and active acquisition, and supports both labeled-only and mixed-supervision learning.

\textbf{Sparse annotation for object detection}. Sparsely annotated object detection (SAOD)~\citesupp{supwu2018soft,supniitani2019partaware,supzhang2020BRL,supwang2021comining,suprambhatla2022sparsely}, deals with the interference of unlabeled labels, which will be considered as hard negatives during the training of detectors. The methods in this field mainly fall into two classes: loss re-weighting and pseudo-labeling. Soft sampling~\citesupp{supwu2018soft} and Background Re-calibration~\citesupp{supzhang2020BRL} trust the judgements of the detector and accordingly adjust the weights of negative proposals. More recent methods draw on a Siamese network~\citesupp{supwang2021comining} or consistency regularization~\citesupp{suprambhatla2022sparsely} to generate pseudo-labels for unlabeled regions.

The proposed box-level active detection framework inevitably incurs the similar challenge. While SAOD methods are validated on randomly down-sampled benchmark datasets, our setting distinguishes itself in prioritizing the annotation of challenging targets while requiring remedies for the easier ones. 
Without specific handling, the performance of detection would be severely interfered. But meantime we can benefit from the prior knowledge of the unlabeled targets: pseudo-labeling that accepts confident model predictions can exactly supplement easier targets with self-supervision. 

\textbf{Mixed types of supervision for object detection}. 
Besides the box-level supervision we focus on in this paper, some recent work also includes additional image-level labels~\citesupp{supbiffi2020many,supzhong2022mixed,supfang2020ehsod} and more types of supervision (\eg scribbles in \citesupp{supren2020omniufo}) to facilitate the detection performance. 
For example, \citesupp{supbiffi2020many,supzhong2022mixed,supfang2020ehsod} utilize weakly labeled datasets and a proportion of fully labeled images.
In comparison, our box-only setting is simpler and more effective.
Take the VOC dataset for example, as presented in Tab.~\ref{tab:voc}, sCOMPAS obtains 73.57\% mAP50 with merely 8.5\% (4K) \emph{boxes}, and then achieves 76.4\% mAP50 with 10.6\% (5K) \emph{boxes}. 
Given access to all \emph{unlabeled images}, mCOMPAS reaches 79.97\%-80.63\% mAP50. 
It clearly surpasses the state-of-the-art 69.4\% mAP50 in \citesupp{supbiffi2020many,supzhong2022mixed,supfang2020ehsod}, where \emph{all image-level labels} and 10\% \emph{fully labeled images} are required.

Mixed types of supervision can also be sampled via active learning.
Based on a pre-trained weakly supervised detector, BiB~\citesupp{supBiB_eccv22} proposes to actively provide full box annotations for images. 
BAOD~\citesupp{suppardo2021baod} progressively adds image-level supervision or full box annotations in each active learning cycle. 
In contrast to their mixed types of supervision and exhaustive annotation protocol, the proposed pipeline is simple and effective in de-redundancy, and is superior in detection performance.

\section{Numerical Results}
Last, we present the exact numerical results used to plot Fig.~1 and Fig.~3 of the main paper and the results of Fig.~\ref{fig:coco54} in this supplementary file.
Tab.~\ref{tab:voc} reports mAP50 and standard deviation (\%) under the VOC-sup and VOC-semi settings. Tab.~\ref{tab:coco} presents results under COCO-sup and COCO-semi settings with the 20K box-level annotation batch size, and Tab.~\ref{tab:coco54} presents the results with a larger batch size of 40K boxes.

\newpage
{\small
\bibliographystylesupp{ieee_fullname}
\bibliographysupp{egbib_sup}
}

\end{document}